\documentclass[10pt,twocolumn,letterpaper]{article}

\usepackage[pagenumbers]{style}

\usepackage{graphicx}
\usepackage{amsmath}
\usepackage{amssymb}
\usepackage{booktabs}
\usepackage{makecell}
\usepackage{bm}
\usepackage{enumitem}
\usepackage{bbding}
\usepackage{tablefootnote}
\usepackage[misc]{ifsym}

\usepackage[pagebackref,breaklinks,colorlinks]{hyperref}
\usepackage[table]{xcolor}
\usepackage{adjustbox}
\usepackage{multirow}
\def\eg{\textit{e.g.,~}}
\def\ie{\textit{i.e.,~}}
\def\comp{\ensuremath\mathop{\scalebox{.6}{$\circ$}}}
\newcommand\smallalign[1]{\begingroup\small
    \setlength{\abovedisplayskip}{0.7em}
    \setlength{\belowdisplayskip}{0.7em}
    \setlength{\abovedisplayshortskip}{0.7em}
    \setlength{\belowdisplayshortskip}{0.7em}
    {#1}\endgroup}
    
\usepackage[capitalize]{cleveref}
\crefname{section}{Sec.}{Secs.}
\Crefname{section}{Section}{Sections}
\Crefname{table}{Table}{Tables}
\crefname{table}{Tab.}{Tabs.}

\def\name{M3I Pre-training}

\newcommand{\newunderbrace}[2]{\begingroup
      \color{blue}
      \underbrace{\color{black}#1}_{\text{#2}}
      \endgroup}
\definecolor{mygray}{gray}{0.3}
\newcommand{\demph}[1]{\textcolor{mygray}{#1}}

\newcommand{\RED}[1]{\textcolor{red}{{#1}}}
\definecolor{defaultcolor}{gray}{.87}
\newcommand{\default}[1]{\textbf{#1}}

\newcommand\blfootnote[1]{%
\begingroup
\renewcommand\thefootnote{}\footnote{#1}%
\addtocounter{footnote}{-1}%
\endgroup
}

\begin{document}

\title{Towards All-in-one Pre-training via Maximizing \\ Multi-modal Mutual Information}

\author{
  Weijie Su$^{1*\dag}$,
  Xizhou Zhu$^{2,4*}$\textsuperscript{\Letter},
  Chenxin Tao$^{3*\dag}$,
  Lewei Lu$^{2}$,
  Bin Li$^{1}$,
  Gao Huang$^{3}$,\\
  Yu Qiao$^{4}$, 
  Xiaogang Wang$^{5,2}$,
  Jie Zhou$^{3}$,
  Jifeng Dai$^{3,4}$\\
$^1$University of Science and Technology of China \ \ \
$^2$SenseTime Research \ \ \
$^3$Tsinghua University\\  
$^4$Shanghai Artificial Intelligence Laboratory \ \ \
$^5$The Chinese University of Hong Kong\\
{\tt\small jackroos@mail.ustc.edu.cn, \{zhuwalter,luotto\}@sensetime.com}\\
{\tt\small tcx20@mails.tsinghua.edu.cn, binli@ustc.edu.cn, \{gaohuang,jzhou,daijifeng\}@tsinghua.edu.cn}\\
{\tt\small qiaoyu@pjlab.org.cn, xgwang@ee.cuhk.edu.hk}}

\maketitle

\begin{abstract}
  \vspace{-0.5em}
  To effectively exploit the potential of large-scale models, various pre-training strategies supported by massive data from different sources are proposed, including supervised pre-training, weakly-supervised pre-training, and self-supervised pre-training. It has been proved that combining multiple pre-training strategies and data from various modalities/sources can greatly boost the training of large-scale models. However, current works adopt a multi-stage pre-training system, where the complex pipeline may increase the uncertainty and instability of the pre-training. It is thus desirable that these strategies can be integrated in a single-stage manner. In this paper, we first propose a general multi-modal mutual information formula as a unified optimization target and demonstrate that all existing approaches are special cases of our framework. Under this unified perspective, we propose an all-in-one single-stage pre-training approach, named \textbf{M}aximizing \textbf{M}ulti-modal \textbf{M}utual \textbf{I}nformation Pre-training (\textbf{\name{}}). Our approach achieves better performance than previous pre-training methods on various vision benchmarks, including ImageNet classification, COCO object detection, LVIS long-tailed object detection, and ADE20k semantic segmentation. Notably, we successfully pre-train a billion-level parameter image backbone and achieve state-of-the-art performance on various benchmarks. Code shall be released at \url{https://github.com/OpenGVLab/M3I-Pretraining}.
\end{abstract}
\blfootnote{$^{*}$ Equal contribution. $^{\dag}$This work is done when Weijie Su and Chenxin Tao are interns at Shanghai Artificial Intelligence Laboratory. \Letter\ Corresponding author.}
\vspace{-1.5em}
\section{Introduction}
\label{sec:intro}
\vspace{-0.5em}

\begin{figure}
    \centering
    \includegraphics[width=\linewidth]{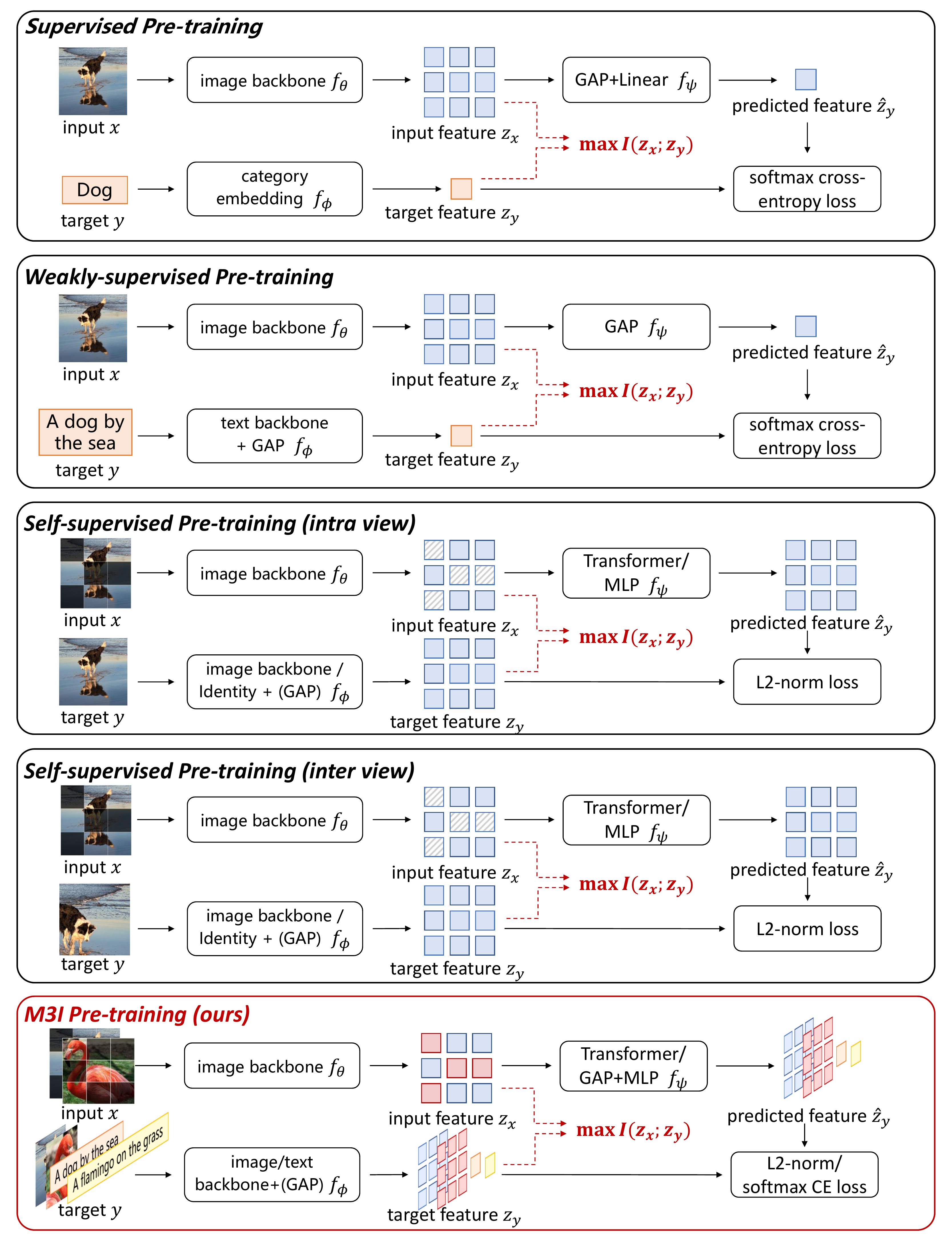}
    \vspace{-2.0em}
    \caption{Comparison between different pre-training paradigms and \name{}. Existing pre-training methods are all optimizing the mutual information between the input and target representations, which can be integrated by \name{}.}
    \label{fig:comparison}
    \vspace{-1.5em}
\end{figure}

In recent years, large-scale pre-trained models~\cite{tan2019efficientnet,radford2021learning,jia2021scaling,bao2021beit,he2022masked,chen2020simple,grill2020bootstrap,zbontar2021barlow} have swept a variety of computer vision tasks with their strong performance.
To adequately train large models with billions of parameters, researchers design various annotation-free self-training tasks and obtain sufficiently large amounts of data from various modalities and sources.
In general, existing large-scale pre-training strategies are mainly divided into three types: supervised learning~\cite{tan2019efficientnet,dai2021coatnet} on pseudo-labeled data (\eg JFT-300M~\cite{xie2020self}), weakly supervised learning~\cite{radford2021learning,jia2021scaling} on web crawling images text pairs (\eg LAION-400M~\cite{schuhmann2021laion}), and self-supervised learning~\cite{chen2020simple,grill2020bootstrap,zbontar2021barlow,bao2021beit,he2022masked} on  unlabeled images. Supported by massive data, all these strategies have their own advantages and have been proven to be effective for large models of different tasks.
In pursuit of stronger representations of large models, some recent approaches~\cite{wang2022image,liu2022swin,wei2022contrastive} combine the advantages of these strategies by directly using different proxy tasks at different stages, significantly pushing the performance boundaries of various vision tasks.

Nevertheless,
the pipeline of these multi-stage pre-training approaches is complex and fragile, which may lead to uncertainty and catastrophic forgetting issues.
Specifically, the final performance is only available after completing the entire multi-stage pre-training pipeline. Due to the lack of effective training monitors in the intermediate stages, it is difficult to locate the problematic training stage when the final performance is poor.
To eliminate this dilemma, it is urgent to develop a single-stage pre-training framework that can take advantage of various supervision signals.
It is natural to raise the following question: \textit{Is it possible to design an all-in-one pre-training method to have all the desired representational properties? }

To this end, we first point out that different single-stage pre-training methods share a unified design principle through a generic pre-training theoretical framework. We further extend this framework to a multi-input multi-target setting so that different pre-training methods can be integrated systematically. In this way, we propose a novel single-stage pre-training method, termed \name{}, that all desired representational properties are combined in a unified framework and trained together in a single stage.

Specifically, we first introduce a generic pre-training theoretical framework that can be instantiated to cover existing mainstream pre-training methods. 
This framework aims to maximize the mutual information between input representation
and target representation,
which can be further derived into a prediction term with a regularization term. 
(1) The prediction term reconstructs training targets 
from the network inputs, which is equivalent to existing well-known pre-training losses by choosing proper forms for the predicted distribution.
(2) The regularization term requires the distribution of the target 
to maintain high entropy to prevent collapse, which is usually implemented implicitly through negative samples or stop-gradient operation. As shown in Fig.~\ref{fig:comparison}, by adopting different forms of input-target paired data
and their representations,
our framework can include existing pre-training approaches and provide possible directions to design an all-in-one pre-training method.

To meet the requirement of large-scale pre-training with various data sources, we further extend our framework to the multi-input multi-target setting, with which we show that multi-task pre-training methods are optimizing a lower bound of the mutual information. 
In addition, we mix two masked views from two different images as the input. The representation of one image is used to reconstruct the same view, while the other image is used to reconstruct a different augmented view. Both representations will predict their corresponding annotated category or paired texts.
In this way, we propose a novel pre-training approach, called \name{}, which can effectively combine the merits of supervised/weakly-supervised/self-supervised pre-training and enables large-scale vision foundation models to benefit from multi-modal/source large-scale data. Our contributions can be summarized as follows:
\begin{itemize}[leftmargin=1em]
    \vspace{-0.5em}
    \item We theoretically demonstrate all existing mainstream pre-training methods share a common optimization objective, \ie maximizing the mutual information between input and target representation. We also show how to instantiate our framework as distinct pre-training methods.    
    \item We propose a novel single-stage pre-training approach called \name\ to gather the benefit of various pre-training supervision signals, via extending our mutual information pre-training framework to a multi-input multi-target setting.
    \item Comprehensive experiments demonstrate the effectiveness of our approach. We successfully pre-train InternImage-H\cite{anonymous2022internimg}, a model with billion-level parameters, and set a new record on basic detection and segmentation tasks, \ie 65.4 box AP on COCO test-dev~\cite{lin2014microsoft}, 62.9 mIoU on ADE20K~\cite{zhou2019semantic}.
\end{itemize}

\section{Related Work}
\label{sec:related}

\noindent\textbf{Supervised Pre-training (SP)} has been the mainstream over a long period of time~\cite{girshick2014rich,he2016deep,chen2017deeplab,dosovitskiy2020image,carion2020end,liu2021swin,liu2022convnet}. Most works adopt image classification on ImageNet~\cite{deng2009imagenet} as the pre-training task, for both ConvNets~\cite{he2016deep,xie2017aggregated,tan2019efficientnet,liu2022convnet} and Transformers~\cite{dosovitskiy2020image,touvron2021training,liu2021swin}. SP has benefited many downstream tasks, including object detection~\cite{girshick2014rich,carion2020end}, semantic semantation~\cite{chen2017deeplab,xiao2018unified}, and video recognition~\cite{bertasius2021space}. Some works have also explored the scaling properties of pre-training datasets~\cite{dosovitskiy2020image,zhai2022scaling} and image backbones~\cite{zhai2022scaling, abnar2021exploring}. Moreover, SP shows that mixing two inputs~\cite{zhang2017mixup,yun2019cutmix} is critical for improving accuracy~\cite{touvron2021training,Touvron2022DeiTIR}, which is rarely explored in other pre-training paradigms. Our proposed framework includes SP as a special case. \name{} can thus preserve the advantage of it in an all-in-one pre-training and surpass its performances on downstream tasks.

\vspace{0.5em}\noindent\textbf{Self-supervised Pre-training (SSP)} becomes popular in recent years~\cite{jaiswal2020survey,zhang2022survey}. It does not require annotations, and thus enables the usage of large-scale unlabeled data. SSP can be divided into two kinds of methods: Intra-view tasks create input and target from the same view, which includes auto-encoder~\cite{hinton2006reducing,vincent2008extracting}, global/dense distillation~\cite{hinton2015distilling,wei2022contrastive} and masked image modeling (MIM)~\cite{bao2021beit,he2022masked,baevski2022data2vec,chen2022context,xie2022simmim}. On the contrary, inter-view tasks adopt different augmented views as input and target, such as dense/global instance discrimination~\cite{wu2018unsupervised, oord2018representation} and siamese image modeling~\cite{anonymous2022siamese}. 
Instance discrimination contains several sub-frameworks, including contrastive learning~\cite{he2020momentum, chen2020simple}, asymmetric networks~\cite{grill2020bootstrap,chen2021exploring} and feature decorrelation~\cite{zbontar2021barlow,bardes2021vicreg}, which are found of similar mechanism~\cite{tian2021understanding,tao2022exploring}.
Some SSP methods have displayed great potential by surpassing SP on downstream tasks by a large margin~\cite{he2020momentum,chen2020simple,he2022masked}. MIM has also been proven to work well with large-scale networks~\cite{he2022masked}. SiameseIM~\cite{anonymous2022siamese} is a inter-view SSP method that better combines semantic alignment with spatial sensitivity. Our method covers different self-supervised pre-training methods in a general framework and seeks to find the most effective setting through extensive experiments. It can combine all the strengths and shows impressive performances.

\vspace{0.5em}\noindent\textbf{Weakly-supervised Pre-training (WSP)} utilizes image-text datasets~\cite{sharma2018conceptual,changpinyo2021conceptual,thomee2016yfcc100m,schuhmann2021laion} or image-hashtag datasets~\cite{mahajan2018exploring,veit2018separating,singh2022revisiting}. These pre-training methods rely on noisy supervision from the Internet and are thus scalable.
For image-hashtag datasets, some works~\cite{mahajan2018exploring,singh2022revisiting} show competitive performances in various transfer-learning settings.
For image-text datasets, earlier works~\cite{Su2020VL-BERT:,lu2019vilbert,chen2020uniter,sun2019videobert,sun2019learning,tan2019lxmert,li2020unicoder,alberti2019fusion,li2019visualbert} mainly focused on learning general representations for visual-linguistic understanding. Recently CLIP~\cite{radford2021learning} and ALIGN\cite{jia2021scaling} demonstrated the effectiveness of image-text pre-training in image recognition. They propose to learn the aligned visual-linguistic representations and achieve impressive image classification accuracy in a zero-shot manner. In our framework, WSP is shown to be a special instance. \name{} leverages the power of WSP to achieve a new height on various downstream tasks.

\vspace{0.5em}\noindent\textbf{Multi-task Pre-training} adopts multiple targets for the same input. This kind of method usually combines text target from WSP and image target from SSP~\cite{yu2022coca,singh2022flava,dong2022maskclip,mu2022slip}. Some works have also explored using both category targets from SP and image targets from SSP~\cite{khosla2020supervised,liang2022supmae}. 
Multi-task pre-training fits well into our framework with the multi-input multi-target extension. Compared with previous works, our method can be successfully applied to large-scale models and displays superior results.

\vspace{0.5em}\noindent\textbf{Multi-stage Pre-training} instead adopts stage-wise pre-training, which focuses on one pre-training target in each stage and reuses the model in the next stage~\cite{peng2022beit,wang2022image,wei2022contrastive}. Multi-stage pre-training also follows the mutual information objective in each stage. However, multi-stage pre-training suffers from a complex pipeline that may increase uncertainty and instability. On the contrary, \name{} combines different supervision signals in a single stage that avoids the problems of multi-stage pre-training.

\section{Method}
\label{sec:theory}
\subsection{Mutual Information for Generic Pre-training}

\setlength{\tabcolsep}{3pt}
\renewcommand{\arraystretch}{1.05}
\begin{table*}[t]
    \centering
    \resizebox{1.0\linewidth}{!}{
    \begin{tabular}{lccccccc}
    \toprule
    Pre-training Method & \makecell[c]{Typical Work} & \makecell[c]{Input\\ Data $x$} & \makecell[c]{Target\\ Data $y$} & \makecell[c]{Input\\ Representation $z_x$} & \makecell[c]{Target\\ Representation $z_y$} & \makecell[c]{Regularization \\ $H\big(p(z_y | t_y)\big)$}  &  \makecell[c]{Distribution \\ Form $\hat{P}$} \\
    \midrule
    \multicolumn{7}{l}{\textit{\demph{Supervised Pre-training :}}} \\
    Image Classification & ViT~\cite{dosovitskiy2020image} & view1 & category & dense feature & category embedding & negative categories & Boltzmann \\
    \midrule
    \multicolumn{7}{l}{\textit{\demph{Weakly-supervised Pre-training :}}} \\
    \makecell[l]{Contrastive Language-\\Image Pre-training} & CLIP~\cite{radford2021learning} & view1 & text & dense feature & text embedding & negative texts & Boltzmann \\
    \midrule
    \multicolumn{7}{l}{\textit{\demph{Self-supervised Pre-training (intra-view) :}}} \\
    Auto-Encoder & - & view1 & view1 & dense feature & dense pixels & - & Gaussian \\
    \RED{$^1$}Dense Distillation & FD\cite{wei2022contrastive},BEiT v2 tokenizer\cite{peng2022beit} & view1 & view1 & dense feature & dense feature & stop gradient & Gaussian \\
    Global Distillation & - & view1 & view1 & dense feature & global feature & stop gradient & Boltzmann \\
    $\text{Masked Image Modeling}_{\text{pixel}}$ & MAE~\cite{he2022masked} & masked view1 & view1 & dense feature & dense pixels & - & Gaussian%
    \vspace{0.3em}\\
    $\RED{^2}\text{Masked Image Modeling}_{\text{feature}}$ & \makecell{data2vec\cite{baevski2022data2vec},MILAN\cite{MILAN2022},\\BEiT\cite{bao2021beit},BEiT v2\cite{peng2022beit}} & masked view1 & view1 & dense feature & dense feature & stop gradient & Gaussian%
    \vspace{0.3em}\\
    $\text{Masked Image Modeling}_{\text{global}}$ & - & masked view1 & view1 & dense feature & global feature & stop gradient & Gaussian \\
    \midrule
    \multicolumn{7}{l}{\textit{\demph{Self-supervised Pre-training (inter-view) :}}} \\
    Novel View Synthesis & - & view2 & view1 & dense feature & dense pixels & - & Gaussian \\
    Dense Instance Discrimination & DenseCL~\cite{wang2021dense} & view2 & view1 & dense feature & dense feature & negative samples & Boltzmann%
    \vspace{0.3em}\\
    \RED{$^3$}Instance Discrimination & \makecell[c]{MoCo\cite{he2020momentum},BYOL\cite{grill2020bootstrap},\\Barlow Twins~\cite{zbontar2021barlow}} & view 2 & view1 & dense feature & global feature & \makecell[c]{negative samples / stop\\ gradient / decorrelation} & \makecell[c]{Boltzmann \\ / Gaussian}%
    \vspace{0.3em}\\
    $\text{Siamese Image Modeling}_{\text{pixel}}$ & - & masked view2 & view1 & dense feature & dense pixels & - & Gaussian \\
    $\text{Siamese Image Modeling}_{\text{feature}}$ & SiameseIM~\cite{anonymous2022siamese} & masked view2 & view1 & dense feature & dense feature & stop gradient & Gaussian \\
    $\text{Siamese Image Modeling}_{\text{global}}$ & MSN~\cite{assran2022masked} & masked view2 & view1 & dense feature & global feature & \makecell[c]{negative samples} & Boltzmann \\
    \bottomrule
    \end{tabular}}
    \vspace{-0.5em}
    \caption{Instances of our mutual information based pre-training framework. Methods that do not have a listed typical work have rarely been explored before as a pre-training method. We only include single-input single-target pre-training methods in this table. 
    \RED{$^1$}Input representation of Dense Distillation can be continuous (FD) or discrete (BEiT v2 tokenizer).
    \RED{$^2$}Target encoder of $\text{Masked Image Modeling}_{\text{feature}}$ can be momentum encoder (data2vec), pre-trained image encoder (MILAN), dVAE (BEiT), or discrete tokenizer distilled from pre-trained image encoder (BEiT v2). 
    \RED{$^3$}Regularization term of Instance Discrimination can be negative samples (MoCo), stop-gradient (BYOL), or decorrelation (Barlow Twins).}
    \vspace{-1.0em}
    \label{tab:comparison}
\end{table*}

The goal of pre-training is to learn representations that can well represent the training samples.
Suppose the training sample is $s$, where $s$ could be image-category pair (supervised), image-text pair (weakly-supervised), or image only (self-supervised). The input training data $x$ and target training data $y$ are extracted from $s$ by some transform operations $(t_x, t_y)$, \ie $x = t_x(s), y = t_x(s)$. The transform operations $t$ is typically data-irrelevant, \eg ``apply image augmentation'', ``get annotated category'' or ``get paired text''. In vision-centric pre-training, the input data $x$ is usually an augmented image, while the target data $y$ is either the annotated category, the paired text, or also an augmented image. $z_x$ and $z_y$ denote the encoded representation for input and target, respectively. Then, the desired pre-training objective can be described as maximizing the conditional mutual information between $z_x$ and $z_y$ given $t_x$ and $t_y$ as (see Appendix for derivation):
\smallalign{\begin{align*}
%\label{equ:data_rep}
    &(s, t_x, t_y) \sim D_\text{train} &&\text{(sampled training sample),}\\
    &x = t_x(s) ~,~ y = t_y(s) &&\text{(extracted training data),}\\
    &z_x \sim p(z_x | x) ~,~ z_y \sim p(z_y | y)  &&\text{(encoded training representation),} 
\end{align*}}%
\vspace{-1em}
\smallalign{\begin{align}
\label{equ:mutual_info}
     I(z_x&; z_y ~|~ t_x, t_y) = \newunderbrace{\mathbb{E}_{p(t_y)} \Big[H\big(p(z_y | t_y)\big) \Big]}{regularization term to avoid collapse}  \nonumber\\
 &-\newunderbrace{\mathbb{E}_{p(s,t_x,t_y,z_x)}\Big[H\big(p(z_y|y)~,~ p(z_y|z_x, t_x,  t_y)\big)\Big]}{(cross-entropy) prediction term for target representation},  
\end{align}}%
where the first term requires high entropy of the target representation, which avoids collapse. The second term requires the posterior distribution $p(z_y|z_x, t_x, t_y)$ to be close to the target distribution $p(z_y|y)$. 

In practice, deterministic neural networks are used for encoding the input representation $z_x$ and target representation $z_y$. On the other hand, the posterior distribution $p(z_y|z_x, t_x, t_y)$ is usually intractable. To alleviate this issue, a common practice is introducing another parameterized distribution $p_\psi(z_y|z_x, t_x, t_y)$ as an approximation, Then, Eq.~\eqref{equ:mutual_info} becomes (see Appendix for derivation):
\smallalign{\begin{align*}
    &z_x = f_\theta(x), ~\quad z_y = f_\phi(y) &&\text{(parameterized representation),}\\
    &\hat{z}_y= f_\psi(z_x, t_x, t_y) &&\text{(prediction of $z_y$ given $z_x$),}\\
    &p_\psi(z_y|z_x, t_x,  t_y) = \hat{P}(z_y~|~\hat{z}_y) &&\text{(approximated distribution),}\\
\end{align*}}%
\vspace{-2em}
\smallalign{\begin{align}
\label{equ:mutual_simple}
    I(z_x; &z_y ~|~ t_x, t_y) = ~\sup_{f_\psi}~ \newunderbrace{\mathbb{E}_{p(t_y)} \Big[H\big(p(z_y(\phi) ~|~ t_y)\big) \Big]}{regularization term to avoid collapse}  \nonumber\\
    &+\newunderbrace{\mathop{\mathbb{E}}_{p(s, t_x, t_y)}\Big[\log \hat{P}\big(z_y(\phi)~|~\hat{z}_y(\theta, \psi)\big)\Big],}{\text{(log-likelihood) prediction term for target representation}}
\end{align}}%
where $\hat{P}$ is the approximated posterior distribution of $z_y$ given the prediction of $\hat{z}_y$. When $z_y$ is continuous and deterministic given $y$, the regularization term becomes intractable~\cite{belghazi2018mine,tishby2000information}. Different mechanisms would be incorporated to avoid representation collapse (see Sec~\ref{sec:conn}). Then, to maximize the mutual information in Eq.~\eqref{equ:mutual_simple}, the training loss can be derived as:
\smallalign{\begin{align}
\label{equ:loss}
    \min_{\theta, \phi, \psi} &\mathop{\mathbb{E}}_{p(s, t_x, t_y)} L(s, t_x, t_y; \theta, \phi, \psi) = -\log \hat{P}\big(z_y(\phi)~|~\hat{z}_y(\theta, \psi)\big), \nonumber\\
    \textit{s.t.~} &\quad\text{non-collapse representation of $z_y$.}
\end{align}}%
Different form of $\hat{P}$ results in different loss, \eg Gaussian and Boltzmann distributions corresponding to L2-norm and Softmax cross-entropy losses, respectively:
\smallalign{\begin{align*}
%\label{equ:g_loss}
    & \hat{P}(z_y~|~\hat{z}_y) \sim \mathcal{N}(\hat{z}_y, \sigma^2 I) \quad\text{(Gaussian distribution)} \\
    \Rightarrow~ &L = -\log \hat{P}(z_y~|~\hat{z}_y) = \frac{1}{2\sigma^2} \|z_y - \hat{z}_y\|^2 + C,\\
% \end{align*}}%
% \vspace{-1em}
% \smallalign{\begin{align*}
%\label{equ:b_loss}
    & \hat{P}(z_y~|~\hat{z}_y) \propto \exp(\hat{z}_y^T z_y / \tau) \quad\text{(Boltzmann distribution)} \\
    \Rightarrow~ &L = -\log \hat{P}(z_y~|~\hat{z}_y) = -\log \frac{\exp(\hat{z}_y^T z_y / \tau)}{\sum_{z_y'} \exp(\hat{z}_y^T z_y' / \tau)},
\end{align*}}%
where $\sigma$ and $\tau$ are the hyper-parameters of Gaussian and Boltzmann distributions, respectively. $C$ is a constant that can be ignored. $z_y'$ iterates over all possible target representations $\{f_\phi(y) ~|~ y = t_y(s) \in D_\text{train}\}$.

Eq.~\eqref{equ:loss} is a generic pre-training loss that can be instantiated into different pre-training paradigms, including supervised, weakly-supervised, and self-supervised pre-training. 
% Tab.~\ref{tab:notation} and 
Tab.~\ref{tab:comparison} demonstrate the actual implementation of different pre-training methods. Different methods incorporate different mechanisms to avoid representation collapse. 

\subsection{Connection with Existing Pre-training Methods}
\label{sec:conn}

%\vspace{0.5em}
\noindent\textbf{Supervised Pre-training (SP)} usually adopts \textit{Image Classification (IC)} as the pre-training task. It takes an augmented image $I$ as input data and the corresponding annotated category $C$ as the target data. The input representation is $z_x = f_\theta(I)$, while the target representation is the category embedding (\eg linear classification weight) $z_y = f_\phi(C)$. The classifier predicts the category based on $z_x$ as $\hat{z}_y= f_\psi(z_x)$. Thus, the pre-training objective is to maximize $I\big(f_\theta(I)~;~ f_\phi(C)\big)$, and the SP loss can be derived as minimizing $L = -\log \hat{P}\big(f_\phi(C) ~|~ f_\psi \comp f_\theta (I)\big)$.
\smallalign{\begin{align*}
%\label{equ:sup}
    &\max~ I\big(f_\theta(I)~;~ f_\phi(C)\big) \nonumber\\
    \Rightarrow~ &\min~ L = -\log \hat{P}\big(f_\phi(C) ~|~ f_\psi \comp f_\theta (I)\big),
\end{align*}}%
where $\hat{P}$ is typically Boltzmann distribution (\ie Softmax cross-entropy loss). This distribution contains negative categories and naturally prevents collapse.
As a mainstream pre-training framework, SP has been proven to be helpful on many downstream tasks over a long period of time~\cite{girshick2014rich,carion2020end,chen2017deeplab,xiao2018unified}. It learns from clean human-annotated data. This helps the model to develop common semantics and converge faster on downstream tasks.

\vspace{0.5em}\noindent\textbf{Weakly-supervised Pre-training (WSP)} usually adopts \textit{Contrastive Language-Image Pre-training (CLIP)}~\cite{radford2021learning,jia2021scaling} as the pre-training task. It takes an augmented image $I$ as input, and the corresponding paired text $T$ as targets. Similar to supervised learning, the pre-training objective is
\smallalign{\begin{align*}
%\label{equ:w_sup}
    &\max~ I\big(f_\theta(I)~;~ f_\phi(T)\big) \nonumber\\
    \Rightarrow~ &\min~ L = -\log \hat{P}\big(f_\phi(T) ~|~ f_\psi \comp f_\theta (I)\big),
\end{align*}}%
where $\hat{P}$ is also Boltzmann distribution, which contains negative samples to prevent the collapse. WSP is able to exploit the massive image-text pairs from the Internet. With the help of image-text alignment, it not only enables many possible new tasks, \eg open-vocabulary recognition~\cite{gu2021open,radford2021learning}, but also greatly boosts the performances of classification and detection tasks in long-tail scenario~\cite{tian2022vl}.

\vspace{0.5em}\noindent\textbf{Self-supervised Pre-training (SSP)} learns representation using images only. Given a sampled training image $I$, the input data is an augmented view of this image $\tilde{I}_x = t_x(I)$, the target data is another augmented view $\tilde{I}_y = t_y(I)$. The pre-training objective is derived from Eq.~\eqref{equ:loss} as
\smallalign{\begin{align*}
%\label{equ:s_sup}
    &\max~ I\big(f_\theta(\tilde{I}_x) ~;~ f_\phi(\tilde{I}_y)\big) \nonumber\\
    \Rightarrow~ &\min~ L = -\log \hat{P}\Big(f_\phi(\tilde{I}_y) ~|~ f_\psi \big(f_\theta (\tilde{I}_x), t_x, t_y\big)\Big),
\end{align*}}%
where $t_x$ and $t_y$ are the input and target augmentations on the sampled image, respectively. Depending on different methods, the target encoder $f_\phi$ could be identity, shared with $f_\theta$ or the Exponential Moving Average (EMA) of $f_\theta$.
$\hat{P}$ is usually Boltzmann or Gaussian distribution. When $\hat{P}$ is Boltzmann (\ie Softmax cross-entropy loss), it aims to differentiate $z_y$ from different training data. When $\hat{P}$ is Gaussian (\ie L2-norm loss), it fits the value of $z_y$. To prevent collapse, ``stop-gradient''~\cite{grill2020bootstrap}, feature-decorrelation~\cite{zbontar2021barlow} and negative samples~\cite{chen2020simple} are considered.

As Tab.~\ref{tab:comparison} illustrated, different choices of data transform operations $(t_x, t_y)$ and target representation type $z_y$ result in different pre-training tasks: (1) For $(t_x, t_y)$, they could be either the the same view (\eg auto-encoder) or different views (\eg instance discrimination~\cite{chen2020simple,grill2020bootstrap,zbontar2021barlow}). $t_x$ could also incorporate an additional mask operation (\eg masked image modeling~\cite{bao2021beit,he2022masked}). (2) For $z_y$, its representation type could be from \{dense pixels, dense feature, global feature\}.

The advantage of SSP methods is that they can utilize large-scale unlabelled data, which facilitates the development of large models. Some SSP methods can already surpass SP on downstream tasks~\cite{he2020momentum,he2022masked,bao2021beit}. Notably, MIM~\cite{he2022masked,bao2021beit} demonstrates great dense localization ability, while SiameseIM~\cite{anonymous2022siamese} can exhibit semantic alignment and spatial sensitivity at the same time.

\begin{figure*}
    \centering
    \includegraphics[width=0.95\textwidth]{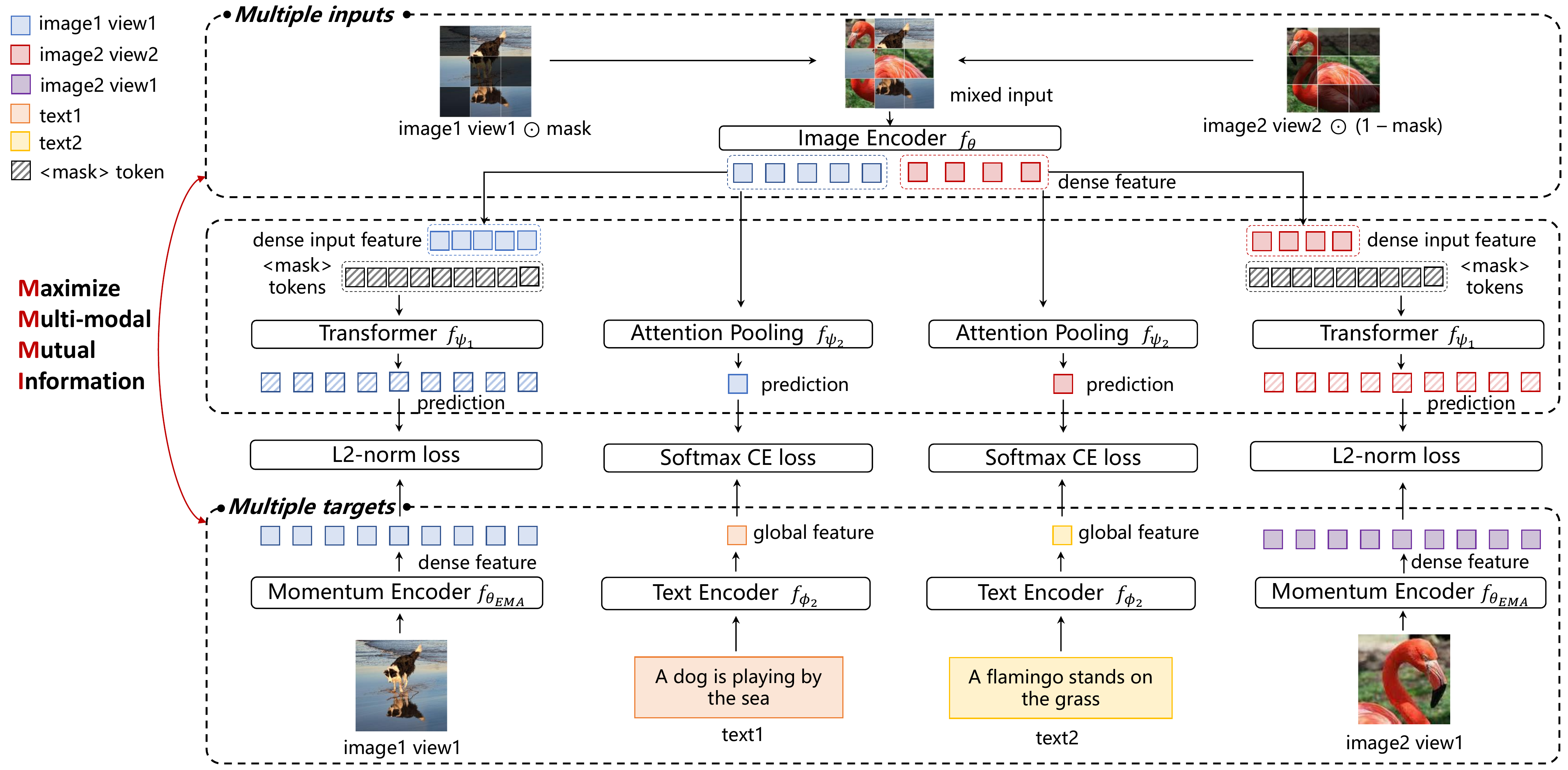}
    \vspace{-0.5em}
    \caption{Overview of \name{}. We mix two views from different images as the inputs. The first image needs to predict the same image view, while the other image needs to predict a different augmented view. Both images need to predict their annotated category or paired text. When predicting image targets, we add position embeddings to decoder inputs so that ``$<$mask$>$'' tokens can be matched with target patches. Following~\cite{anonymous2022siamese}, position embeddings are computed with respect to the left top origin of the input views.}
    \label{fig:my_label}
    \vspace{-1.0em}
\end{figure*}

\subsection{Multi-input Multi-target Pre-training}
Based on previous analysis, we can see that different pre-training tasks possess their own strengths. Naturally, we would like to maintain all these properties in one pre-training approach. 
For this purpose, we extend our framework to a multi-input multi-target setting.

Suppose the set of $N$ multiple inputs and $M$ multiple targets are $X = \{x_i\}_{i=1}^N$ and $Y = \{y_j\}_{j=1}^M$, respectively. We use $\bm{t}_{x}$ and $\bm{t}_{y}$ to indicate the sets of transforms of inputs and targets. 

In practice, most methods choose to optimize the objectives of different types of targets separately. In this case, we can split the targets $Y$ into $K$ non-overlapping groups and encode different groups independently as $Y_m \cap Y_{n\neq m}=\varnothing, \cup_{k=1}^KY_k= Y$. With this modification, we show that the mutual information in Eq.~\eqref{equ:mutual_simple} can be bounded by (see Appendix for derivation):
\smallalign{\begin{align*}
    &(\bm{s}, \bm{t}_x, \bm{t}_y, X, Y) \sim D_\text{train} &&\text{\footnotesize(sample inputs and targets)} \\
    &\bm{z}_x = f_\theta\big(X = \{x_i\}_{i=1}^N\big) &&\text{\footnotesize(encode multiple inputs jointly)}\\
    &\bm{z}_y^k = f_{\phi_k}(Y_k), ~ Y_k=\{y_{k_j}\}_{j=1}^{M_k} &&\text{\footnotesize(encode multiple targets separately)}\\
    &{\bm{\hat{z}}_y}^k= f_{\psi_k}(\bm{z}_x, \bm{t}_x, \bm{t}_y) &&\text{\footnotesize(predict multiple targets separately)}
\end{align*}}%
\vspace{-1em}
\smallalign{\begin{align}
\label{equ:multi_mutual}
    I\big(\bm{z}_x;& \{\bm{z}_y^k\}_{k=1}^{K} | \bm{t}_x, \bm{t}_y\big) \geq \sup_{\{f_{\psi_k}\}_{k=1}^{K}} \newunderbrace{\mathop{\mathbb{E}}_{p(\bm{t}_y)} \bigg[H\Big(p\big(\{\bm{z}_y^k\}_{k=1}^{K} | \bm{t}_y\big)\Big) \bigg]}{regularization term to avoid collapse} \nonumber\\
    &+\newunderbrace{\sum_{k=1}^K~ \mathop{\mathbb{E}}_{p(\bm{s}, \bm{t}_x, \bm{t}_y)}\Big[\log \hat{P}_k\big(\bm{z}_y^k~|~\bm{\hat{z}}_y^k\big)\Big]}{(log-likelihood) prediction term for target representation}, \nonumber\\
    \Rightarrow~ &L(\bm{s}, \bm{t}_x, \bm{t}_y) = \sum_{k=1}^K -\log \hat{P}_k\big(\bm{z}_y^k(\phi_k)~|~\bm{\hat{z}}_y^k(\theta, \psi_k)\big),
\end{align}}%
where $k$ is the group index, $M_k$ is the number of targets in $k^{\text{th}}$ group, and $\hat{P}_k$ is the approximated distribution for each target group.
Each prediction term corresponds to the objective of a target group. This implies that optimizing target objectives independently is equivalent to optimizing a lower bound of the mutual information.

\vspace{0.5em}\noindent\textbf{Multi-input Pre-training ($N = M$)} uses multiple inputs with one target for each input $X = \{x_i\}_{i=1}^N,  Y = \{y_i\}_{i=1}^N$,

where $y_i$ is the corresponding sampled target of $x_i$.
multi-input pre-training is widely used in supervised pre-training (typically $N = 2$), where different images are mixed through Mixup~\cite{zhang2017mixup} or CutMix~\cite{yun2019cutmix}. It has proven to be critical for improving accuracy and providing a more stable generalization ability. However, in other pre-training paradigms, a single input is usually adopted. The lack of multiple inputs may hinder better model performance, and also lead to inconsistency between different pre-training paradigms, which hampers the pre-training combination.

\vspace{0.5em}\noindent\textbf{Multi-target Pre-training ($N = 1$)} only uses multiple targets for the same input as $X = x, Y = \{y_i\}_{i=1}^M$.

Some previous works have explored the use of multiple targets~\cite{yu2022coca,singh2022flava,khosla2020supervised,liang2022supmae}. One line of research tries to combine weakly-supervised pre-training with specific forms of self-supervised pre-training, such as MaskCLIP~\cite{dong2022maskclip} and FLAVA~\cite{singh2022flava}. Another line studies the combination of supervised pre-training and self-supervised pre-training, such as SupCon~\cite{khosla2020supervised} or SupMAE~\cite{liang2022supmae}. These methods display the effectiveness of multiple targets.

\subsection{\name{}}
\label{subsec:theory_mix}
With the help of our mutual information framework, we are able to systematically integrate different pre-trainings into a whole, which we name as \name{}. 
It has two inputs and four targets, combining self-supervised and supervised / weakly-supervised pre-training as
\smallalign{\begin{align*}
   X = \{\tilde{I}_{x_i}, \tilde{I}_{x_j}\}, ~\quad Y = \{\tilde{I}_{y_i}, \tilde{I}_{y_j}, T_i, T_j\},
\end{align*}}%
where $\tilde{I}_{x_i}, \tilde{I}_{x_j}$ are the augmented input views of two different sampled images $I_i, I_j$, and $\tilde{I}_{y_i}, \tilde{I}_{y_j}$ are the corresponding augmented target views. $T_i, T_j$ denotes the corresponding annotated category (supervised) or paired text (weakly-supervised) for each image.

\vspace{0.5em}\noindent\textbf{Input Encoder $f_\theta$} first mixes the input views with a randomized binary mask $m$ as $\tilde{I}_\text{mix} = m \odot \tilde{I}_{x_i} + (1-m) \odot \tilde{I}_{x_j}$,

where $\odot$ is the element-wise product, $m$ shares the same shape as inputs. Then, the input representation is encoded by an image backbone (\eg ViT~\cite{dosovitskiy2020image}) as $f_\theta(\tilde{I}_\text{mix})$.
In order to make this mix strategy compatible with existing pre-training tasks like Masked Image Modeling (MIM) and Image Classification (IC), we split the mask $m$ into patches with $p \times p$ size. All pixels in the same patch will be masked or unmasked together. For example, $p = 16$ is by default used for MIM~\cite{he2022masked, bao2021beit}. Note that the widely used Mixup~\cite{zhang2017mixup} and CutMix~\cite{yun2019cutmix} are generally incompatible with MIM.

\vspace{0.5em}\noindent\textbf{Target Encoder $f_\phi$} is responsible for producing the target representations. 
For image targets $\tilde{I}_{y_i}, \tilde{I}_{y_j}$, we use the momentum input image backbone as the encoder to generate dense target features. For category targets (supervised) or text targets (weakly-supervised) $T_i, T_j$, we use a category embedding or text backbone that is jointly trained during pre-training. Notice that because of the multiple inputs, we can adopt both intra-view and inter-view self-supervised predictions: the first image $i$ is asked to predict the same view (\ie $\tilde{I}_{x_i} = \tilde{I}_{y_i}$), and the other image $j$ instead needs to predict a different augmented view (\ie $\tilde{I}_{x_j} \neq \tilde{I}_{y_j}$).

\vspace{0.5em}\noindent\textbf{Input-to-Target Decoder $f_\psi$} predicts the target representations from the input. For simplicity, we use the separate loss form in Eq.~\eqref{equ:multi_mutual} to predict each target separately. We adopt Transformer~\cite{vaswani2017attention} layers to predict the dense representations for image targets, and an attention pooling layer~\cite{chen2022context} followed by a linear projection to predict the category embedding (supervised) or text embedding (weakly-supervised).

\section{Experiment}
\label{sec:exp}
\noindent\textbf{Implementation Details.~}  We utilize InternImage-H~\cite{anonymous2022internimg} as image encoder in Sec~\ref{sec:exp_large_model} for large-scale model pre-training and ViT-B/16~\cite{dosovitskiy2020image} as that in other experiments for ablation study and fair comparison. For image-text dataset (\eg YFCC-15M~\cite{thomee2016yfcc100m}), a 12-layer Transformer (with the same network architecture as BERT-Base~\cite{devlin2018bert}) is utilized as text target encoder. For image classification dataset (\eg ImageNet-1k~\cite{deng2009imagenet}), we directly use the linear classifier weight as category embedding target. We employ 4-layer Transformer as decoder for image representation target, and Attention Pooling as that for category embedding or text global feature. Please see Appendix for detailed pre-training hyper-parameters.

\subsection{Pre-training of 1B Image Backbone}
\label{sec:exp_large_model}

\setlength{\tabcolsep}{3pt}
\renewcommand{\arraystretch}{1.2}
\begin{table*}[t]
    \centering
    \small
    \resizebox{1.0\linewidth}{!}{
        \begin{tabular}{cllllccccc}
        \Xhline{2\arrayrulewidth}
         Pre-training Approach & Model & Pipeline & Public Data & Private Data & \makecell{ImageNet\\val} & \makecell{COCO\\test-dev} & \makecell{LVIS\\minival}& \makecell{ADE20k\\val}\\
        \hline
        \multirow{2}{*}{\name{}} & \multirow{2}{*}{InternImage-H~\cite{anonymous2022internimg} (1B)} & \multirow{2}{*}{Single Stage: \name{}} & \multirow{2}{*}{\makecell[l]{427M image-text\\ 15M image-category}} & \multirow{2}{*}{\qquad-} &\multirow{2}{*}{89.2} & \multirow{2}{*}{\textbf{65.4}}  & \multirow{2}{*}{\textbf{62.5}}  &  \multirow{2}{*}{\textbf{62.9}} \\
        & \\
        \hline
        \multirow{2}{*}{\cite{liu2022swin}} & \multirow{2}{*}{SwinV2-G (3B)} & Stage 1: Masked Image Modeling$_{\text{pixel}}$ & \multirow{2}{*}{\makecell[l]{15M image-category}} & \multirow{2}{*}{55M image-category} & \multirow{2}{*}{89.2} & \multirow{2}{*}{63.1} & \multirow{2}{*}{-} &  \multirow{2}{*}{59.9}  \\
        & & Stage 2: Image Classification \\
        \hline
        \multirow{3}{*}{\cite{wang2022image}} & \multirow{3}{*}{BEiT-3 (2B)} & Stage 1: CLIP & \multirow{3}{*}{\makecell[l]{21M image-text\\ 15M image-category}} & \multirow{3}{*}{400M image-text} & \multirow{3}{*}{\textbf{89.6}} & \multirow{3}{*}{63.7} & \multirow{3}{*}{-} & \multirow{3}{*}{62.8} \\
        & & Stage 2: Dense Distillation &  \\
        & & Stage 3: Masked Data Modeling & &  \\
        \hline
        \multirow{3}{*}{\cite{wei2022contrastive}} & \multirow{3}{*}{SwinV2-G (3B)} & Stage 1: Masked Image Modeling$_{\text{pixel}}$ & \multirow{3}{*}{\makecell[l]{15M image-category}} & \multirow{3}{*}{55M image-category} & \multirow{3}{*}{89.4} & \multirow{3}{*}{64.2} & \multirow{3}{*}{-} & \multirow{3}{*}{61.4} \\
        & & Stage 2: Image Classification \\
        & & Stage 3: Dense Distillation \\
        \hline
        \multirow{2}{*}{$^\dag$previous best} & & & & & \multirow{2}{*}{89.1\RED{$^a$}} & \multirow{2}{*}{64.5\RED{$^b$}} & \multirow{2}{*}{59.8\RED{$^c$}} & \multirow{2}{*}{60.8\RED{$^d$}} \\
        \\
        \Xhline{2\arrayrulewidth}
        \end{tabular}
    }
    \vspace{-0.5em}
    \caption{Comparision of \name{} with existing large model pre-training methods for visual recognition. Top1 Accuracy, AP$^{\text{box}}$, AP$^{\text{box}}$, mIoU are reported on ImageNet validation set, COCO test-dev set, LVIS minival set (to avoid  data contamination following ~\cite{zhang2022glipv2}), and ADE20k validation set, respectively. We achieve state-of-the-art performance on object detection and semantic segmentation tasks. \name{} also demonstrate very competitive classification performance with only public datasets, while all other methods utilize large-scale private data (WIT-400M~\cite{radford2021learning} is used in ~\cite{wang2022image}, ImageNet-22k-ext~\cite{liu2022swin} is used in \cite{liu2022swin,wei2022contrastive}), which is strong correlated with the task of image classification. $^\dag$We also list previous best results on these tasks with only public training data for comparision. Results reference: \RED{$^a$}MOAT~\cite{yang2022moat}, \RED{$^b$}Group DETR v2~\cite{chen2022group}, \RED{$^c$}GLIPv2~\cite{zhang2022glipv2}, \RED{$^d$}Mask DINO~\cite{li2022mask}.} 
    \vspace{-1.0em}
    \label{tab:exp_sota}
\end{table*}

\vspace{0.5em}\noindent\textbf{Settings.~} We employ InternImage-H~\cite{anonymous2022internimg} (a ConvNet-based image backbone with 1B parameters) as image encoder. The network is pre-trained for 30 epochs on 427M public image-text pairs (LAION400M~\cite{schuhmann2021laion}, YFCC-15M~\cite{kalkowski2015real}, CC12M~\cite{changpinyo2021conceptual}) and 15M public image-category pairs (ImageNet-22k~\cite{deng2009imagenet}). We report the transfer performance on ImageNet~\cite{deng2009imagenet}, COCO~\cite{lin2014microsoft}, LVIS~\cite{gupta2019lvis}, and ADE20k~\cite{zhou2017scene} benchmarks.

\vspace{0.5em}\noindent\textbf{Results and Discussions.~} As shown in Tab.~\ref{tab:exp_sota}, all previous large model pre-training approaches adopt a complicated multi-stage training pipeline. Instead, our \name{} is a simple yet effective single-stage pre-training paradigm. It achieves state-of-the-art performance on COCO object detection, LVIS long-tailed object detection,  and ADE20k semantic segmentation. Very competitive performance is achieved on ImageNet classification. It validates the effectiveness of our approach. Besides, \name{} only employs pubic datasets and exhibits superior transfer performance while all other approaches include private datasets in their pre-training.

Different from SwinV2~\cite{liu2022swin}, BEiT-3~\cite{wang2022image} and FD~\cite{wei2022contrastive}, \name{} is an all-in-one single-stage training paradigm which brings the following advantages: 1) \textit{Simplicity.} \name{}\ could make use of all available supervision signals and training data in a single-stage pre-training. In contrast, both \cite{liu2022swin, wang2022image} incorporate redundant multi-stage pre-training pipelines. \cite{liu2022swin} uses the same training data in multiple pre-training stages but with different supervision signals. \cite{wang2022image} picks the pre-trained model in the previous pre-training stage as the target network for the next pre-training stage.  2) \textit{Avoiding Catastrophic Forgetting}. As shown in Tab.~\ref{tab:exp_sota}, \cite{liu2022swin,wang2022image,wei2022contrastive} all consist of multiple pre-training stages. The networks are expected to learn different representational attributes in different pre-training stages. However, due to the existence of catastrophic forgetting~\cite{french1999catastrophic}, attributes learned in the previous pre-training stage may be forgotten in the next pre-training stage.
Our \name{} naturally avoids the catastrophic forgetting issue by learning different representational attributes simultaneously in one-stage pre-training.

\setlength{\tabcolsep}{2pt}
\renewcommand{\arraystretch}{1.1}
\begin{table}[t]
    \centering
    \small
    \resizebox{1.0\linewidth}{!}{
        \begin{tabular}{clcccccc}
        \Xhline{2\arrayrulewidth}
        & Pre-training Method & \makecell[c]{Input\\ Data $x$} & \makecell[c]{Target\\ Representation $z_y$} & \makecell{ImageNet\\Top1} & \makecell{COCO\\AP$^{\text{box}}$} \\
        \hline
        \multicolumn{6}{l}{\textit{\demph{Self-supervised Pre-training (intra-view)}}} \\
        (a) & Auto-Encoder      & view1         & dense pixels         &  77.5 & 0.0$^{\dag}$ \\
        (b) & Dense Distillation & view1         & dense feature  & 78.8 & 32.4 \\
        (c) & Global Distillation         & view1           & global feature & 77.1 & 27.9\\
        (d) & Masked Image Modeling$_{\text{pixel}}$          & masked view1  & dense pixels   & 83.1 & 46.8 \\
        \default{(e)} & \default{Masked Image Modeling$_{\text{feat}}$}           & \default{masked view1}  & \default{dense feature}  & \default{83.3} & \default{47.4} \\
        (f) & Masked Image Modeling$_{\text{gloal}}$ & masked view1 & global feature & 83.2 & 47.5\\
        \hline
        \multicolumn{6}{l}{\textit{\demph{Self-supervised Pre-training (inter-view)}}} \\
        (g) & Novel View Synthesis                 & view2        & dense pixels     & 78.8 & 33.0 \\     
        (h) & Dense Instance Discrimination             & view2         & dense feature  & 83.2 & 50.1 \\
        (i) & Instance Discrimination                           & view2       & global feature & 83.0 & 46.4  \\
        (j) & Siamese Image Modeling$_{\text{pixel}}$    & masked view2  & dense pixels   & 78.9 & 38.1 \\
        \default{(k)} & \default{Siamese Image Modeling$_{\text{feat}}$}     & \default{masked view2}  & \default{dense feature}  & \default{83.7} & \default{49.8}  \\
        (l) & Instance Discrimination$_{\text{mask}}$   & masked view2  & global feature & 82.9 & 46.2 \\
        \Xhline{2\arrayrulewidth}
        \end{tabular}
    }
    \vspace{-0.5em}
    \caption{Ablation study on different self-supervised pre-training methods under our framework. $^\dag$ denotes no convergence in fine-tuning. }
    \vspace{-1em}
    \label{tab:exp_ablation_selfsup}
\end{table}

\setlength{\tabcolsep}{4pt}
\begin{table}
    \centering
    \small
    \resizebox{1.0\linewidth}{!}{
        \begin{tabular}{lccccc}
        \Xhline{2\arrayrulewidth}
        \multirow{2}{*}{Pre-training Method} & \makecell{ImageNet} & \makecell{COCO} & \multicolumn{2}{c}{LVIS} & \makecell{ADE20k} \\
        & \makecell{Top1} & \makecell{AP$^{\text{box}}$} & AP$^{\text{box}}$ & AP$_{\text{rare}}^{\text{box}}$ & \makecell{mIoU} \\
        \hline
        Image Classification              \textbf{}& 81.8 & 46.6 & 33.0 & 25.5 & 45.1 \\
        Best Intra-view SSP    & 83.3 & 47.4 & 31.2 & 21.9 & 40.1 \\
        Best Inter-view SSP   & 83.7 & 49.8 & 35.2 & 26.9 & 47.7\\
        \hline
        \multicolumn{6\textbf{}}{l}{\textit{\demph{Ours}}} \\
        \name{}~w/o mix  & 83.7 & 50.3 & 36.6 & 27.2 & 48.7  \\
        \default{\name{}} & \default{83.9} & \default{50.8} & \default{37.5} & \default{29.6} & \default{49.0} \\
        \Xhline{2\arrayrulewidth}
        \end{tabular}
    }
    \vspace{-0.5em}
    \caption{Ablation study of multi-input multi-target pre-training.}
    \vspace{-1em}
    \label{tab:exp_ablation_multi}
\end{table}

\setlength{\tabcolsep}{4pt}
\renewcommand{\arraystretch}{1.2}
\begin{table*}
    \centering
    \small
    \resizebox{1.0\linewidth}{!}{
        \begin{tabular}{lc|lllc|rc}
        \Xhline{2\arrayrulewidth}
        \multirow{2}{*}{Task}& \multirow{2}{*}{Metric} & & \multicolumn{2}{c}{ImageNet Pre-train} & & \multicolumn{2}{c}{YFCC Pre-train} \\
             &                           &\makecell[c]{SSP (intra-view)}&\makecell[c]{SSP (inter-view)}&\makecell[c]{SP} & M3I (ImageNet) &\makecell[c]{WSP} & M3I (YFCC) \\ 
        \hline
        ImageNet w/o Fine-tuning & Top1 acc.       & \makecell[c]{$\times$}  & \makecell[c]{$\times$} & \default{83.8} (DeiT-III) & 83.3 & $^\dag$37.6 (CLIP) & $^\dag$39.1~~ \\
        ImageNet Linear Classification & Top1 acc. & 79.5 (iBOT)             & 78.0 (SiameseIM)       & \default{83.8} (DeiT-III) & \default{83.8} & 66.5 (CLIP)     & 72.3 \\
        ImageNet Fine-tuning & Top1 acc.           & \default{84.2} (data2vec)         & 84.1 (SiameseIM)       & 83.8 (DeiT-III) & \default{84.2} &  80.5 (CLIP)    & 83.7 \\
        \hline                                                                                                                 
        COCO & AP$^{\text{box}}$                   & 51.6 (MAE)              & 52.1 (SiameseIM)       &  47.6 (Sup.)    & \default{52.2} & \makecell[c]{-} & 51.9 \\
        \hline                                                                                                                 
        \multirow{2}{*}{LVIS} &  AP$^{\text{box}}$ & 40.1 (MAE)              & 40.5 (SiameseIM)       & 37.2 (Sup.)     & 40.6 & \makecell[c]{-} & \default{40.8}\\
        &  AP$^{\text{box}}_{\text{rare}}$         & 38.1 (MAE)              & 38.1 (SiameseIM)       & \makecell[c]{-} & 38.2 & \makecell[c]{-} & \default{38.4} \\
        \hline                                                                                                                 
        ADE20k &  mIoU                             & 50.0 (iBOT)             & 51.1 (SiameseIM)       & 49.3 (DeiT-III) & \default{51.3} & \makecell[c]{-} & \default{51.3} \\
        \Xhline{2\arrayrulewidth}
        \end{tabular}
    }
    \vspace{-0.5em}
    \caption{System-level comparison with SoTA supervised, weakly-supervised, self-supervised pre-training methods (\ie DeiT-III~\cite{Touvron2022DeiTIR}, CLIP~\cite{radford2021learning}, data2vec~\cite{baevski2022data2vec}, MAE~\cite{he2022masked}, iBOT~\cite{zhou2021ibot}and SiameseIM~\cite{anonymous2022siamese}). The results of CLIP are obtained from \cite{mu2022slip} for a fair comparison on YFCC-15M. The results of ``Sup.'' (refers to supervised pre-training) are obtained from ~\cite{li2022exploring}. All methods adopt ViT-B/16 as the image backbone for a fair comparison. The fine-tuning schedule is 100 epochs for ImageNet, COCO, LVIS, and ADE20k. $\times$ denotes do not support. $^\dag$ ``ImageNet w/o Fine-tuning'' for weakly-supervised pre-training corresponds to the zero-shot transfer setting.}
    \vspace{-1.0em}
    \label{tab:exp_main}
\end{table*}

\subsection{Ablation Study}
\label{sec:exp_ablation}

\vspace{0.5em}\noindent\textbf{Ablation Settings.~} We utilize ViT-B/16 as the image backbone for the ablation study. The pre-training schedule is set to 400 epochs on ImageNet-1k. Different pre-training methods are evaluated by the transfer performance on ImageNet-1k classification, COCO detection, LVIS detection, and ADE20k segmentation. The fine-tuning schedule is 100 epochs for ImageNet-1k. For other datasets, fine-tuning with 25 epochs is adopted.

\vspace{0.5em}\noindent\textbf{Ablation on Self-supervised Pre-training (SSP).~} As Tab.~\ref{tab:comparison} shows, the mutual information framework proposes 12 forms of SSP, some of which have not been explored as pre-training before. We compare these 12 types of SSP in Tab.~\ref{tab:exp_ablation_selfsup}. Based on the experiment results, We analyze three key factors in these approaches:

\noindent\emph{1) Masked Input or Full Input.~} Masked input is critical for both intra-view and inter-view pre-training. The performances of Tab.~\ref{tab:exp_ablation_selfsup} (d-f,j-l) (masked) are always better or on par with Tab.~\ref{tab:exp_ablation_selfsup} (a-c,g-i)(full). The comparison for intra-view pre-training is consistent with previous studies~\cite{he2022masked}, implying that masking operation can greatly boost the model's performance. We observe that the gap for inter-view pre-training becomes smaller. The reason may be that predicting another view constitutes a more challenging task, and reduce the information redundancy to some extent.

\noindent\emph{2) Target Representation.~} The dense feature works best under almost all settings. Compared to the global feature target, the dense feature target enables the spatial discrimination ability of the network. Thus, as shown in Tab.~\ref{tab:exp_ablation_selfsup} (kl) and Tab.~\ref{tab:exp_ablation_selfsup} (hi), it achieves much better performance on COCO. On the other hand, compared to dense pixels, dense features represent the target in high-level semantic space and thus bring semantic alignment capacity. For example, Tab.~\ref{tab:exp_ablation_selfsup} (k) surpasses Tab.~\ref{tab:exp_ablation_selfsup} (j) by a large margin both in ImageNet (+4.8 points) and COCO (+11.7 points).

\noindent\emph{3) Intra-view or Inter-view.~}
The choice of intra-view or inter-view pre-training depends on whether the input data is masked or not. If full input is adopted, inter-view generally performs better than intra-view, as shown in Tab.~\ref{tab:exp_ablation_selfsup} (a-c, g-i). We conjecture that recovering the same view is too easy, and may not be suitable for pre-training.
On the other hand, if masked input is employed, both intra-view and inter-view can find a setting with good performance, \eg Tab.~\ref{tab:exp_ablation_selfsup} (ek).

\vspace{0.5em}\noindent\textbf{Ablation on Multi-input Multi-target Pre-training.~} After we have determined the best training setting for SSP with intra-view and inter-view, we can now combine different pre-training methods into an all-in-one approach with our multi-input multi-target framework. Tab.~\ref{tab:exp_ablation_multi} demonstrates the comparison between \name{} and single-input single-target pre-training methods. Here we only consider pre-training on supervised pre-training for simplicity.

We first compare multi-target pre-training, \ie{} \name{} w/o mix, with single-input single-target pre-training methods. It's shown that \name{} w/o mix can obtain superior or comparable results on all tasks, especially on LVIS (+1.4 points) and ADE20k (+1.0 points) benchmarks. We note that even though some pre-training methods may not perform well on some tasks, the combination is still effective to improve upon all single-input single-target methods. This is because different pre-training methods focus on different representational properties. For example, Image Classification pre-training brings better semantic information. This leads to high results on LVIS and ADE20k datasets, where long-tail classes pose high demand for semantic understanding. Intra-view SSP instead excels on spatial sensitivity and delivers good performance on the COCO dataset. \name{} w/o mix demonstrates the benefits of these methods. Our final \name{} further adopts multiple inputs to better combine these pre-training methods. Experiments show that it achieves better performances on all tasks.

\subsection{System-level Comparision with Other Methods}
\label{sec:exp_main}

We compare \name{} with previous methods using the same ViT-B/16~\cite{dosovitskiy2020image} image backbone in Tab.~\ref{tab:exp_main}. We pre-train our model for 1600 epochs and finetune it for 100 epochs on ImageNet~\cite{deng2009imagenet}, COCO~\cite{lin2014microsoft}, LVIS~\cite{gupta2019lvis} and ADE20k~\cite{zhou2017scene} datasets. We also report the results on ImageNet without finetuning and with the linear protocol. We further validate our method on YFCC-15M image-text dataset~\cite{kalkowski2015real}. For a fair comparison, the pre-training iterations are kept the same with ImageNet pre-training.

Tab.~\ref{tab:exp_main} shows that different pre-training methods possess different advantages. SP learns semantic alignment well and can already deliver good performance on ImageNet without further finetuning. WSP can enable zero-shot transfer learning, which can not be achieved through other pre-training methods. SSP presents better localization ability and is vital for dense prediction tasks. \name{} is able to achieve comparable results with the best of previous methods on all these tasks. This indicates that \name{} can maintain all these desired properties through a single-stage pre-training.

\section{Conclusion}
\label{sec:conclusion}
Modern large-scale networks rely on combining different pre-training methods to effectively utilize massive data, which suffers from the multi-stage pre-training practice. To derive a single-stage pre-training, we proposed a generic pre-training framework that unifies mainstream pre-training approaches. We further extended the framework to a multi-input multi-target setting, which shows that previous multi-task pre-training methods are actually optimizing a lower bound of the mutual information. Finally, we proposed an all-in-one pre-training method, \name{}. \name{} surpasses previous pre-training methods in various transfer-learning settings. 

\noindent\textbf{Limitations.} We focused on vision-centric pre-training. The proposed framework can be applied to other domains, like natural language processing or visual-linguistic tasks. We expect to explore other domains in future work.

\vspace{-1.0em}
\paragraph{Acknowledgments} The work is partially supported by the National Natural Science Foundation of China under grants No.U19B2044, No.61836011 and No.62022048.

{\small
\bibliographystyle{ieee_fullname}
\bibliography{egbib}

\begin{thebibliography}{10}\itemsep=-1pt

\bibitem{abnar2021exploring}
Samira Abnar, Mostafa Dehghani, Behnam Neyshabur, and Hanie Sedghi.
\newblock Exploring the limits of large scale pre-training.
\newblock {\em arXiv preprint arXiv:2110.02095}, 2021.

\bibitem{alberti2019fusion}
Chris Alberti, Jeffrey Ling, Michael Collins, and David Reitter.
\newblock Fusion of detected objects in text for visual question answering.
\newblock {\em arXiv preprint arXiv:1908.05054}, 2019.

\bibitem{assran2022masked}
Mahmoud Assran, Mathilde Caron, Ishan Misra, Piotr Bojanowski, Florian Bordes,
  Pascal Vincent, Armand Joulin, Mike Rabbat, and Nicolas Ballas.
\newblock Masked siamese networks for label-efficient learning.
\newblock In {\em ECCV}, pages 456--473. Springer, 2022.

\bibitem{baevski2022data2vec}
Alexei Baevski, Wei-Ning Hsu, Qiantong Xu, Arun Babu, Jiatao Gu, and Michael
  Auli.
\newblock Data2vec: A general framework for self-supervised learning in speech,
  vision and language.
\newblock {\em arXiv preprint arXiv:2202.03555}, 2022.

\bibitem{bao2021beit}
Hangbo Bao, Li Dong, Songhao Piao, and Furu Wei.
\newblock Beit: Bert pre-training of image transformers.
\newblock In {\em ICLR}, 2021.

\bibitem{bardes2021vicreg}
Adrien Bardes, Jean Ponce, and Yann LeCun.
\newblock Vicreg: Variance-invariance-covariance regularization for
  self-supervised learning.
\newblock {\em arXiv preprint arXiv:2105.04906}, 2021.

\bibitem{belghazi2018mine}
Mohamed~Ishmael Belghazi, Aristide Baratin, Sai Rajeswar, Sherjil Ozair, Yoshua
  Bengio, Aaron Courville, and R~Devon Hjelm.
\newblock Mine: mutual information neural estimation.
\newblock {\em arXiv preprint arXiv:1801.04062}, 2018.

\bibitem{bertasius2021space}
Gedas Bertasius, Heng Wang, and Lorenzo Torresani.
\newblock Is space-time attention all you need for video understanding?
\newblock In {\em ICML}, volume~2, page~4, 2021.

\bibitem{carion2020end}
Nicolas Carion, Francisco Massa, Gabriel Synnaeve, Nicolas Usunier, Alexander
  Kirillov, and Sergey Zagoruyko.
\newblock End-to-end object detection with transformers.
\newblock In {\em ECCV}, pages 213--229. Springer, 2020.

\bibitem{changpinyo2021conceptual}
Soravit Changpinyo, Piyush Sharma, Nan Ding, and Radu Soricut.
\newblock Conceptual 12m: Pushing web-scale image-text pre-training to
  recognize long-tail visual concepts.
\newblock In {\em CVPR}, 2021.

\bibitem{chen2017deeplab}
Liang-Chieh Chen, George Papandreou, Iasonas Kokkinos, Kevin Murphy, and Alan~L
  Yuille.
\newblock Deeplab: Semantic image segmentation with deep convolutional nets,
  atrous convolution, and fully connected crfs.
\newblock {\em TPAMI}, 40(4):834--848, 2017.

\bibitem{chen2022group}
Qiang Chen, Jian Wang, Chuchu Han, Shan Zhang, Zexian Li, Xiaokang Chen, Jiahui
  Chen, Xiaodi Wang, Shuming Han, Gang Zhang, et~al.
\newblock Group detr v2: Strong object detector with encoder-decoder
  pretraining.
\newblock {\em arXiv preprint arXiv:2211.03594}, 2022.

\bibitem{chen2020simple}
Ting Chen, Simon Kornblith, Mohammad Norouzi, and Geoffrey Hinton.
\newblock A simple framework for contrastive learning of visual
  representations.
\newblock In {\em ICML}, pages 1597--1607. PMLR, 2020.

\bibitem{chen2022context}
Xiaokang Chen, Mingyu Ding, Xiaodi Wang, Ying Xin, Shentong Mo, Yunhao Wang,
  Shumin Han, Ping Luo, Gang Zeng, and Jingdong Wang.
\newblock Context autoencoder for self-supervised representation learning.
\newblock {\em arXiv preprint arXiv:2202.03026}, 2022.

\bibitem{chen2021exploring}
Xinlei Chen and Kaiming He.
\newblock Exploring simple siamese representation learning.
\newblock In {\em CVPR}, 2021.

\bibitem{chen2020uniter}
Yen-Chun Chen, Linjie Li, Licheng Yu, Ahmed~El Kholy, Faisal Ahmed, Zhe Gan, Yu
  Cheng, and Jingjing Liu.
\newblock Uniter: Universal image-text representation learning.
\newblock In {\em ECCV}, 2020.

\bibitem{chen2022vitadapter}
Zhe Chen, Yuchen Duan, Wenhai Wang, Junjun He, Tong Lu, Jifeng Dai, and Yu
  Qiao.
\newblock Vision transformer adapter for dense predictions.
\newblock {\em arXiv preprint arXiv:2205.08534}, 2022.

\bibitem{cheng2021masked}
Bowen Cheng, Ishan Misra, Alexander~G Schwing, Alexander Kirillov, and Rohit
  Girdhar.
\newblock Masked-attention mask transformer for universal image segmentation.
\newblock {\em arXiv preprint arXiv:2112.01527}, 2021.

\bibitem{mmseg2020}
MMSegmentation Contributors.
\newblock {MMSegmentation}: Openmmlab semantic segmentation toolbox and
  benchmark.
\newblock \url{https://github.com/open-mmlab/mmsegmentation}, 2020.

\bibitem{dai2021coatnet}
Zihang Dai, Hanxiao Liu, Quoc~V Le, and Mingxing Tan.
\newblock Coatnet: Marrying convolution and attention for all data sizes.
\newblock {\em NeurIPS}, 34:3965--3977, 2021.

\bibitem{deng2009imagenet}
Jia Deng, Wei Dong, Richard Socher, Li-Jia Li, Kai Li, and Li Fei-Fei.
\newblock Imagenet: A large-scale hierarchical image database.
\newblock In {\em CVPR}, 2009.

\bibitem{devlin2018bert}
Jacob Devlin, Ming-Wei Chang, Kenton Lee, and Kristina Toutanova.
\newblock Bert: Pre-training of deep bidirectional transformers for language
  understanding.
\newblock {\em arXiv preprint arXiv:1810.04805}, 2018.

\bibitem{dong2022maskclip}
Xiaoyi Dong, Yinglin Zheng, Jianmin Bao, Ting Zhang, Dongdong Chen, Hao Yang,
  Ming Zeng, Weiming Zhang, Lu Yuan, Dong Chen, et~al.
\newblock Maskclip: Masked self-distillation advances contrastive
  language-image pretraining.
\newblock {\em arXiv preprint arXiv:2208.12262}, 2022.

\bibitem{dosovitskiy2020image}
Alexey Dosovitskiy, Lucas Beyer, Alexander Kolesnikov, Dirk Weissenborn,
  Xiaohua Zhai, Thomas Unterthiner, Mostafa Dehghani, Matthias Minderer, Georg
  Heigold, Sylvain Gelly, et~al.
\newblock An image is worth 16x16 words: Transformers for image recognition at
  scale.
\newblock In {\em ICLR}, 2020.

\bibitem{french1999catastrophic}
Robert~M French.
\newblock Catastrophic forgetting in connectionist networks.
\newblock {\em Trends in cognitive sciences}, 3(4):128--135, 1999.

\bibitem{girshick2014rich}
Ross Girshick, Jeff Donahue, Trevor Darrell, and Jitendra Malik.
\newblock Rich feature hierarchies for accurate object detection and semantic
  segmentation.
\newblock In {\em CVPR}, pages 580--587, 2014.

\bibitem{grill2020bootstrap}
Jean-Bastien Grill, Florian Strub, Florent Altch{\'e}, Corentin Tallec, Pierre
  Richemond, Elena Buchatskaya, Carl Doersch, Bernardo Avila~Pires, Zhaohan
  Guo, Mohammad Gheshlaghi~Azar, et~al.
\newblock Bootstrap your own latent-a new approach to self-supervised learning.
\newblock 2020.

\bibitem{gu2021open}
Xiuye Gu, Tsung-Yi Lin, Weicheng Kuo, and Yin Cui.
\newblock Open-vocabulary object detection via vision and language knowledge
  distillation.
\newblock {\em arXiv preprint arXiv:2104.13921}, 2021.

\bibitem{gupta2019lvis}
Agrim Gupta, Piotr Dollar, and Ross Girshick.
\newblock Lvis: A dataset for large vocabulary instance segmentation.
\newblock In {\em CVPR}, pages 5356--5364, 2019.

\bibitem{he2022masked}
Kaiming He, Xinlei Chen, Saining Xie, Yanghao Li, Piotr Doll{\'a}r, and Ross
  Girshick.
\newblock Masked autoencoders are scalable vision learners.
\newblock In {\em CVPR}, 2022.

\bibitem{he2020momentum}
Kaiming He, Haoqi Fan, Yuxin Wu, Saining Xie, and Ross Girshick.
\newblock Momentum contrast for unsupervised visual representation learning.
\newblock In {\em CVPR}, 2020.

\bibitem{he2016deep}
Kaiming He, Xiangyu Zhang, Shaoqing Ren, and Jian Sun.
\newblock Deep residual learning for image recognition.
\newblock In {\em CVPR}, pages 770--778, 2016.

\bibitem{hinton2015distilling}
Geoffrey Hinton, Oriol Vinyals, Jeff Dean, et~al.
\newblock Distilling the knowledge in a neural network.
\newblock {\em arXiv preprint arXiv:1503.02531}, 2015.

\bibitem{hinton2006reducing}
Geoffrey~E Hinton and Ruslan~R Salakhutdinov.
\newblock Reducing the dimensionality of data with neural networks.
\newblock {\em science}, 313(5786):504--507, 2006.

\bibitem{MILAN2022}
Zejiang Hou, Fei Sun, Yen-Kuang Chen, Yuan Xie, and Sun-Yuan Kung.
\newblock Milan: Masked image pretraining on language assisted representation.
\newblock {\em arXiv preprint arXiv:2208.06049}, 2022.

\bibitem{jaiswal2020survey}
Ashish Jaiswal, Ashwin~Ramesh Babu, Mohammad~Zaki Zadeh, Debapriya Banerjee,
  and Fillia Makedon.
\newblock A survey on contrastive self-supervised learning.
\newblock {\em Technologies}, 9(1):2, 2020.

\bibitem{jia2021scaling}
Chao Jia, Yinfei Yang, Ye Xia, Yi-Ting Chen, Zarana Parekh, Hieu Pham, Quoc Le,
  Yun-Hsuan Sung, Zhen Li, and Tom Duerig.
\newblock Scaling up visual and vision-language representation learning with
  noisy text supervision.
\newblock In {\em ICML}, pages 4904--4916. PMLR, 2021.

\bibitem{kalkowski2015real}
Sebastian Kalkowski, Christian Schulze, Andreas Dengel, and Damian Borth.
\newblock Real-time analysis and visualization of the yfcc100m dataset.
\newblock In {\em Proceedings of the 2015 workshop on community-organized
  multimodal mining: opportunities for novel solutions}, pages 25--30, 2015.

\bibitem{khosla2020supervised}
Prannay Khosla, Piotr Teterwak, Chen Wang, Aaron Sarna, Yonglong Tian, Phillip
  Isola, Aaron Maschinot, Ce Liu, and Dilip Krishnan.
\newblock Supervised contrastive learning.
\newblock {\em NeurIPS}, 33:18661--18673, 2020.

\bibitem{li2022mask}
Feng Li, Hao Zhang, Shilong Liu, Lei Zhang, Lionel~M Ni, Heung-Yeung Shum,
  et~al.
\newblock Mask dino: Towards a unified transformer-based framework for object
  detection and segmentation.
\newblock {\em arXiv preprint arXiv:2206.02777}, 2022.

\bibitem{li2020unicoder}
Gen Li, Nan Duan, Yuejian Fang, Ming Gong, and Daxin Jiang.
\newblock Unicoder-vl: A universal encoder for vision and language by
  cross-modal pre-training.
\newblock In {\em AAAI}, 2020.

\bibitem{li2019visualbert}
Liunian~Harold Li, Mark Yatskar, Da Yin, Cho-Jui Hsieh, and Kai-Wei Chang.
\newblock Visualbert: A simple and performant baseline for vision and language.
\newblock {\em arXiv preprint arXiv:1908.03557}, 2019.

\bibitem{li2022exploring}
Yanghao Li, Hanzi Mao, Ross Girshick, and Kaiming He.
\newblock Exploring plain vision transformer backbones for object detection.
\newblock {\em arXiv preprint arXiv:2203.16527}, 2022.

\bibitem{liang2022supmae}
Feng Liang, Yangguang Li, and Diana Marculescu.
\newblock Supmae: Supervised masked autoencoders are efficient vision learners.
\newblock {\em arXiv preprint arXiv:2205.14540}, 2022.

\bibitem{liang2022cbnet}
Tingting Liang, Xiaojie Chu, Yudong Liu, Yongtao Wang, Zhi Tang, Wei Chu,
  Jingdong Chen, and Haibin Ling.
\newblock Cbnet: A composite backbone network architecture for object
  detection.
\newblock {\em IEEE TIP}, 2022.

\bibitem{lin2014microsoft}
Tsung-Yi Lin, Michael Maire, Serge Belongie, James Hays, Pietro Perona, Deva
  Ramanan, Piotr Doll{\'a}r, and C~Lawrence Zitnick.
\newblock Microsoft coco: Common objects in context.
\newblock In {\em ECCV}, pages 740--755. Springer, 2014.

\bibitem{liu2022swin}
Ze Liu, Han Hu, Yutong Lin, Zhuliang Yao, Zhenda Xie, Yixuan Wei, Jia Ning, Yue
  Cao, Zheng Zhang, Li Dong, et~al.
\newblock Swin transformer v2: Scaling up capacity and resolution.
\newblock In {\em CVPR}, pages 12009--12019, 2022.

\bibitem{liu2021swin}
Ze Liu, Yutong Lin, Yue Cao, Han Hu, Yixuan Wei, Zheng Zhang, Stephen Lin, and
  Baining Guo.
\newblock Swin transformer: Hierarchical vision transformer using shifted
  windows.
\newblock In {\em ICCV}, 2021.

\bibitem{liu2022convnet}
Zhuang Liu, Hanzi Mao, Chao-Yuan Wu, Christoph Feichtenhofer, Trevor Darrell,
  and Saining Xie.
\newblock A convnet for the 2020s.
\newblock In {\em CVPR}, 2022.

\bibitem{lu2019vilbert}
Jiasen Lu, Dhruv Batra, Devi Parikh, and Stefan Lee.
\newblock Vilbert: Pretraining task-agnostic visiolinguistic representations
  for vision-and-language tasks.
\newblock 2019.

\bibitem{mahajan2018exploring}
Dhruv Mahajan, Ross Girshick, Vignesh Ramanathan, Kaiming He, Manohar Paluri,
  Yixuan Li, Ashwin Bharambe, and Laurens Van Der~Maaten.
\newblock Exploring the limits of weakly supervised pretraining.
\newblock In {\em ECCV}, pages 181--196, 2018.

\bibitem{mu2022slip}
Norman Mu, Alexander Kirillov, David Wagner, and Saining Xie.
\newblock Slip: Self-supervision meets language-image pre-training.
\newblock In {\em ECCV}, 2022.

\bibitem{oord2018representation}
Aaron van~den Oord, Yazhe Li, and Oriol Vinyals.
\newblock Representation learning with contrastive predictive coding.
\newblock {\em arXiv preprint arXiv:1807.03748}, 2018.

\bibitem{peng2022beit}
Zhiliang Peng, Li Dong, Hangbo Bao, Qixiang Ye, and Furu Wei.
\newblock Beit v2: Masked image modeling with vector-quantized visual
  tokenizers.
\newblock {\em arXiv preprint arXiv:2208.06366}, 2022.

\bibitem{radford2021learning}
Alec Radford, Jong~Wook Kim, Chris Hallacy, Aditya Ramesh, Gabriel Goh,
  Sandhini Agarwal, Girish Sastry, Amanda Askell, Pamela Mishkin, Jack Clark,
  et~al.
\newblock Learning transferable visual models from natural language
  supervision.
\newblock In {\em ICML}, pages 8748--8763. PMLR, 2021.

\bibitem{schuhmann2021laion}
Christoph Schuhmann, Richard Vencu, Romain Beaumont, Robert Kaczmarczyk,
  Clayton Mullis, Aarush Katta, Theo Coombes, Jenia Jitsev, and Aran
  Komatsuzaki.
\newblock Laion-400m: Open dataset of clip-filtered 400 million image-text
  pairs.
\newblock {\em arXiv preprint arXiv:2111.02114}, 2021.

\bibitem{shao2019objects365}
Shuai Shao, Zeming Li, Tianyuan Zhang, Chao Peng, Gang Yu, Xiangyu Zhang, Jing
  Li, and Jian Sun.
\newblock Objects365: A large-scale, high-quality dataset for object detection.
\newblock In {\em ICCV}, pages 8430--8439, 2019.

\bibitem{sharma2018conceptual}
Piyush Sharma, Nan Ding, Sebastian Goodman, and Radu Soricut.
\newblock Conceptual captions: A cleaned, hypernymed, image alt-text dataset
  for automatic image captioning.
\newblock In {\em ACL}, pages 2556--2565, 2018.

\bibitem{singh2022flava}
Amanpreet Singh, Ronghang Hu, Vedanuj Goswami, Guillaume Couairon, Wojciech
  Galuba, Marcus Rohrbach, and Douwe Kiela.
\newblock Flava: A foundational language and vision alignment model.
\newblock In {\em CVPR}, pages 15638--15650, 2022.

\bibitem{singh2022revisiting}
Mannat Singh, Laura Gustafson, Aaron Adcock, Vinicius de Freitas~Reis, Bugra
  Gedik, Raj~Prateek Kosaraju, Dhruv Mahajan, Ross Girshick, Piotr Doll{\'a}r,
  and Laurens van~der Maaten.
\newblock Revisiting weakly supervised pre-training of visual perception
  models.
\newblock In {\em CVPR}, pages 804--814, 2022.

\bibitem{Su2020VL-BERT:}
Weijie Su, Xizhou Zhu, Yue Cao, Bin Li, Lewei Lu, Furu Wei, and Jifeng Dai.
\newblock Vl-bert: Pre-training of generic visual-linguistic representations.
\newblock In {\em ICLR}, 2020.

\bibitem{sun2019learning}
Chen Sun, Fabien Baradel, Kevin Murphy, and Cordelia Schmid.
\newblock Learning video representations using contrastive bidirectional
  transformer.
\newblock {\em arXiv preprint arXiv:1906.05743}, 2019.

\bibitem{sun2019videobert}
Chen Sun, Austin Myers, Carl Vondrick, Kevin Murphy, and Cordelia Schmid.
\newblock Videobert: A joint model for video and language representation
  learning.
\newblock In {\em CVPR}, 2019.

\bibitem{tan2019lxmert}
Hao Tan and Mohit Bansal.
\newblock Lxmert: Learning cross-modality encoder representations from
  transformers.
\newblock In {\em EMNLP}, 2019.

\bibitem{tan2019efficientnet}
Mingxing Tan and Quoc Le.
\newblock Efficientnet: Rethinking model scaling for convolutional neural
  networks.
\newblock In {\em ICML}, pages 6105--6114. PMLR, 2019.

\bibitem{tao2022exploring}
Chenxin Tao, Honghui Wang, Xizhou Zhu, Jiahua Dong, Shiji Song, Gao Huang, and
  Jifeng Dai.
\newblock Exploring the equivalence of siamese self-supervised learning via a
  unified gradient framework.
\newblock In {\em CVPR}, 2022.

\bibitem{anonymous2022siamese}
Chenxin Tao, Xizhou Zhu, Weijie Su, Gao Huang, Bin Li, Jie Zhou, Yu Qiao,
  Xiaogang Wang, and Jifeng Dai.
\newblock Siamese image modeling for self-supervised vision representation
  learning.
\newblock {\em arXiv preprint arXiv:2206.01204}, 2022.

\bibitem{thomee2016yfcc100m}
Bart Thomee, David~A Shamma, Gerald Friedland, Benjamin Elizalde, Karl Ni,
  Douglas Poland, Damian Borth, and Li-Jia Li.
\newblock Yfcc100m: The new data in multimedia research.
\newblock {\em Communications of the ACM}, 59(2):64--73, 2016.

\bibitem{tian2022vl}
Changyao Tian, Wenhai Wang, Xizhou Zhu, Jifeng Dai, and Yu Qiao.
\newblock Vl-ltr: Learning class-wise visual-linguistic representation for
  long-tailed visual recognition.
\newblock In {\em ECCV}, pages 73--91. Springer, 2022.

\bibitem{tian2021understanding}
Yuandong Tian, Xinlei Chen, and Surya Ganguli.
\newblock Understanding self-supervised learning dynamics without contrastive
  pairs.
\newblock In {\em ICML}, pages 10268--10278. PMLR, 2021.

\bibitem{tishby2000information}
Naftali Tishby, Fernando~C Pereira, and William Bialek.
\newblock The information bottleneck method.
\newblock {\em arXiv preprint physics/0004057}, 2000.

\bibitem{touvron2021training}
Hugo Touvron, Matthieu Cord, Matthijs Douze, Francisco Massa, Alexandre
  Sablayrolles, and Herv{\'e} J{\'e}gou.
\newblock Training data-efficient image transformers \& distillation through
  attention.
\newblock In {\em ICML}, pages 10347--10357. PMLR, 2021.

\bibitem{Touvron2022DeiTIR}
Hugo Touvron, Matthieu Cord, and Herve Jegou.
\newblock Deit iii: Revenge of the vit.
\newblock {\em arXiv preprint arXiv:2204.07118}, 2022.

\bibitem{vaswani2017attention}
Ashish Vaswani, Noam Shazeer, Niki Parmar, Jakob Uszkoreit, Llion Jones,
  Aidan~N Gomez, {\L}ukasz Kaiser, and Illia Polosukhin.
\newblock Attention is all you need.
\newblock {\em NeurIPS}, 30, 2017.

\bibitem{veit2018separating}
Andreas Veit, Maximilian Nickel, Serge Belongie, and Laurens van~der Maaten.
\newblock Separating self-expression and visual content in hashtag supervision.
\newblock In {\em CVPR}, pages 5919--5927, 2018.

\bibitem{vincent2008extracting}
Pascal Vincent, Hugo Larochelle, Yoshua Bengio, and Pierre-Antoine Manzagol.
\newblock Extracting and composing robust features with denoising autoencoders.
\newblock In {\em ICML}, pages 1096--1103, 2008.

\bibitem{wang2022image}
Wenhui Wang, Hangbo Bao, Li Dong, Johan Bjorck, Zhiliang Peng, Qiang Liu, Kriti
  Aggarwal, Owais~Khan Mohammed, Saksham Singhal, Subhojit Som, et~al.
\newblock Image as a foreign language: Beit pretraining for all vision and
  vision-language tasks.
\newblock {\em arXiv preprint arXiv:2208.10442}, 2022.

\bibitem{anonymous2022internimg}
Wenhai Wang, Jifeng Dai, Zhe Chen, Zhenhang Huang, Zhiqi Li, Xizhou Zhu,
  Xiaowei Hu, Tong Lu, Lewei Lu, Hongsheng Li, Xiaogang Wang, and Yu Qiao.
\newblock Internimage: Exploring large-scale vision foundation models with
  deformable convolutions.
\newblock {\em arXiv preprint arXiv:2211.05778}, 2022.

\bibitem{wang2021dense}
Xinlong Wang, Rufeng Zhang, Chunhua Shen, Tao Kong, and Lei Li.
\newblock Dense contrastive learning for self-supervised visual pre-training.
\newblock In {\em CVPR}, 2021.

\bibitem{wei2022contrastive}
Yixuan Wei, Han Hu, Zhenda Xie, Zheng Zhang, Yue Cao, Jianmin Bao, Dong Chen,
  and Baining Guo.
\newblock Contrastive learning rivals masked image modeling in fine-tuning via
  feature distillation.
\newblock {\em arXiv preprint arXiv:2205.14141}, 2022.

\bibitem{wu2018unsupervised}
Zhirong Wu, Yuanjun Xiong, Stella~X Yu, and Dahua Lin.
\newblock Unsupervised feature learning via non-parametric instance
  discrimination.
\newblock In {\em CVPR}, 2018.

\bibitem{xiao2018unified}
Tete Xiao, Yingcheng Liu, Bolei Zhou, Yuning Jiang, and Jian Sun.
\newblock Unified perceptual parsing for scene understanding.
\newblock In {\em ECCV}, pages 418--434, 2018.

\bibitem{xie2020self}
Qizhe Xie, Minh-Thang Luong, Eduard Hovy, and Quoc~V Le.
\newblock Self-training with noisy student improves imagenet classification.
\newblock In {\em CVPR}, pages 10687--10698, 2020.

\bibitem{xie2017aggregated}
Saining Xie, Ross Girshick, Piotr Doll{\'a}r, Zhuowen Tu, and Kaiming He.
\newblock Aggregated residual transformations for deep neural networks.
\newblock In {\em CVPR}, pages 1492--1500, 2017.

\bibitem{xie2022simmim}
Zhenda Xie, Zheng Zhang, Yue Cao, Yutong Lin, Jianmin Bao, Zhuliang Yao, Qi
  Dai, and Han Hu.
\newblock Simmim: A simple framework for masked image modeling.
\newblock In {\em CVPR}, pages 9653--9663, 2022.

\bibitem{yang2022moat}
Chenglin Yang, Siyuan Qiao, Qihang Yu, Xiaoding Yuan, Yukun Zhu, Alan Yuille,
  Hartwig Adam, and Liang-Chieh Chen.
\newblock Moat: Alternating mobile convolution and attention brings strong
  vision models.
\newblock {\em arXiv preprint arXiv:2210.01820}, 2022.

\bibitem{yu2022coca}
Jiahui Yu, Zirui Wang, Vijay Vasudevan, Legg Yeung, Mojtaba Seyedhosseini, and
  Yonghui Wu.
\newblock Coca: Contrastive captioners are image-text foundation models.
\newblock {\em arXiv preprint arXiv:2205.01917}, 2022.

\bibitem{yun2019cutmix}
Sangdoo Yun, Dongyoon Han, Seong~Joon Oh, Sanghyuk Chun, Junsuk Choe, and
  Youngjoon Yoo.
\newblock Cutmix: Regularization strategy to train strong classifiers with
  localizable features.
\newblock In {\em ICCV}, pages 6023--6032, 2019.

\bibitem{zbontar2021barlow}
Jure Zbontar, Li Jing, Ishan Misra, Yann LeCun, and St{\'e}phane Deny.
\newblock Barlow twins: Self-supervised learning via redundancy reduction.
\newblock In {\em ICML}, pages 12310--12320. PMLR, 2021.

\bibitem{zhai2022scaling}
Xiaohua Zhai, Alexander Kolesnikov, Neil Houlsby, and Lucas Beyer.
\newblock Scaling vision transformers.
\newblock In {\em CVPR}, 2022.

\bibitem{zhang2022survey}
Chaoning Zhang, Chenshuang Zhang, Junha Song, John Seon~Keun Yi, Kang Zhang,
  and In~So Kweon.
\newblock A survey on masked autoencoder for self-supervised learning in vision
  and beyond.
\newblock {\em arXiv preprint arXiv:2208.00173}, 2022.

\bibitem{zhang2017mixup}
Hongyi Zhang, Moustapha Cisse, Yann~N Dauphin, and David Lopez-Paz.
\newblock mixup: Beyond empirical risk minimization.
\newblock {\em arXiv preprint arXiv:1710.09412}, 2017.

\bibitem{zhang2022dino}
Hao Zhang, Feng Li, Shilong Liu, Lei Zhang, Hang Su, Jun Zhu, Lionel~M Ni, and
  Heung-Yeung Shum.
\newblock Dino: Detr with improved denoising anchor boxes for end-to-end object
  detection.
\newblock {\em arXiv preprint arXiv:2203.03605}, 2022.

\bibitem{zhang2022glipv2}
Haotian Zhang, Pengchuan Zhang, Xiaowei Hu, Yen-Chun Chen, Liunian~Harold Li,
  Xiyang Dai, Lijuan Wang, Lu Yuan, Jenq-Neng Hwang, and Jianfeng Gao.
\newblock Glipv2: Unifying localization and vision-language understanding.
\newblock {\em arXiv preprint arXiv:2206.05836}, 2022.

\bibitem{zhou2017scene}
Bolei Zhou, Hang Zhao, Xavier Puig, Sanja Fidler, Adela Barriuso, and Antonio
  Torralba.
\newblock Scene parsing through ade20k dataset.
\newblock In {\em CVPR}, pages 633--641, 2017.

\bibitem{zhou2019semantic}
Bolei Zhou, Hang Zhao, Xavier Puig, Tete Xiao, Sanja Fidler, Adela Barriuso,
  and Antonio Torralba.
\newblock Semantic understanding of scenes through the ade20k dataset.
\newblock {\em International Journal of Computer Vision}, 127(3):302--321,
  2019.

\bibitem{zhou2021ibot}
Jinghao Zhou, Chen Wei, Huiyu Wang, Wei Shen, Cihang Xie, Alan Yuille, and Tao
  Kong.
\newblock ibot: Image bert pre-training with online tokenizer.
\newblock {\em International Conference on Learning Representations (ICLR)},
  2022.

\end{thebibliography}
}

\newpage
\appendix
\section{Derivation for Mutual Information Framework}
This section describes the detailed derivation for our mutual information framework. For clarity, we list the notations in Tab.~\ref{tab:notation}.
%\newcolumntype{L}[1]{>{\raggedright\arraybackslash}p{#1}}
\setlength{\tabcolsep}{2pt}
\renewcommand{\arraystretch}{1.05}
\begin{table*}[ht!]
    \centering
    \resizebox{\linewidth}{!}{
    \begin{tabular}{llrrr}
    \toprule
    \multirow{2}{*}{Notation} & \multirow{2}{*}{Meaning} & \multicolumn{3}{c}{Typical Choices in Vision-centric Pre-training Paradigms} \\
    \cmidrule{3-5}
     & & Supervised & Weakly-supervised & Self-supervised \\
    \midrule
    $s$ & \textbf{training sample} from the training dataset & image-category pair & image-text pair & image only \\
    \midrule
    $t_x$ & \textbf{input transform operation} applied to the sample $s$ & apply image augmentation & apply image augmentation & apply image augmentation \\
    $t_y$ & \textbf{target transform operation} applied to the sample $s$& get annotated category & get paired text & apply image augmentation%
    \vspace{0.5em}\\
    $\bm{t}_x$ & \textbf{the set of input transform operations} applied to the sample $s$ & - & - & - \\
    $\bm{t}_y$ & \textbf{the set of target transform operations} applied to the sample $s$& - & - & - \\
    \midrule
    $x = t_x(s)$ & \textbf{input data} for the network training & augmented image & augmented image & augmented image \\
    $y = t_y(s)$ & \textbf{target data} for the network training & annotated category & paired text & augmented image%
    \vspace{0.5em}\\
    $\{x_i\}_{i=1}^N = \bm{t}_x(s)$ & \textbf{multiple inputs} for the network training & - & - & - \\
    $\{y_j\}_{j=1}^M = \bm{t}_y(s)$ & \textbf{multiple targets} for the network training & - & - & - \\
    $Y_k = \{y_{k_j}\}_{j=1}^{M_k}$ & \textbf{the $k^{\text{th}}$ group of targets} & - & - & - \\
    \midrule
    $z_x = f_\theta(x)$ & \textbf{input representation} from the input encoder $f_\theta$ & image embedding & image embedding & image embedding \\
    $z_y = f_\phi(y)$ & \textbf{target representation} from the target encoder $f_\phi$ & category embedding & text embedding & image embedding \\
    $\hat{z}_y = f_\psi(z_x, t_x, t_y)$ & \textbf{target prediction} from the decoder $f_\psi$ & predicted embedding & predicted embedding & predicted embedding%
    \vspace{0.5em}\\
    $\bm{z}_x = f_\theta(\{x_i\}_{i=1}^N)$ & \textbf{input representation} from the input encoder $f_\theta$ & - & - & - \\
    $\bm{z}_y^k = f_{\phi_k}(Y^k)$ & \textbf{the $k^{\text{th}}$ group target representation} from the target encoder $f_{\phi_k}$ & - & - & - \\
    $\bm{\hat{z}}_y^k = f_{\psi_k}(\bm{z}_x, t_x, t_y)$ & \textbf{the $k^{\text{th}}$ group target prediction} from the decoder $f_{\psi_k}$ & - & - & - \\
    \midrule
    $\hat{P}(z_y | \hat{z}_y)$ & \textbf{approximated target posterior} given the prediction $\hat{z}_y$ & Boltzmann & Boltzmann & Boltzmann~/~Gaussian%
    \vspace{0.5em}\\
    $\hat{P}_k(\bm{z}_y^k | \bm{\hat{z}}_y^k)$ & \textbf{approximated target posterior} given the prediction $\bm{\hat{z}}_y^k$ & - & - & - \\
    \midrule
    $H\big(p(z_y | t_y)\big)$ & \textbf{regularization term} to avoid representation collapse of $z_y$ & negative categories & negative texts & \makecell[r]{negative samples / stop \\gradient / decorrelation}%
    \vspace{0.5em}\\
    $H\Big(p\big(\{\bm{z}_y^k\}_{k=1}^{K} | \bm{t}_y\big)\Big)$ & \textbf{regularization term} to avoid representation collapse of $\{\bm{z}_y^k\}_{k=1}^{K}$ & - & - & - \\
    \bottomrule
    \end{tabular}}
    \caption{Notation used in this paper. For single-input single-target pre-training, we also list the typical choices in different pre-training paradigm for each notation.}
    \label{tab:notation}
\end{table*}

\subsection{Single-input Sinle-target Pre-training}
We start with the basic form of single-input single-target pre-training. The desired objective is to maximize the conditional mutual information between the input representation $z_x$ and the target representation $z_y$ given the input transform $t_x$ and target transform $t_y$:
\begin{equation}
\small
    \max I(z_x;z_y~|~t_x,t_y).
\end{equation}
According to the definition of conditional mutual information, we have
\begin{align}
\label{equ:append_mutual_info}
\small
    &I(z_x;z_y~|~t_x,t_y) \nonumber\\
    &= \begin{aligned}[t]
        \int p(t_x, t_y)\int\bigg[&p(z_x,z_y~|~t_x,t_y)\cdot \\
        &\log\frac{p(z_y~|~z_x,t_x,t_y)}{p(z_y~|~t_x,t_y)}\bigg]dz_xdz_ydt_xdt_y \nonumber
    \end{aligned}\\
    &= \begin{aligned}[t]
        \int p(t_x, t_y)\int \bigg[&p(z_x,z_y~|~s,t_x,t_y)p(s~|~t_x,t_y)\cdot\\
        &\log\frac{p(z_y~|~z_x,t_x,t_y)}{p(z_y~|~t_x,t_y)}\bigg]dz_xdz_ydt_xdt_yds \nonumber
    \end{aligned}\\
    &= \begin{aligned}[t]
        \int p(s,t_x, t_y)&\int \bigg[p(z_x~|~x)p(z_y~|~y)\cdot\\
        &\quad\log\frac{p(z_y~|~z_x,t_x,t_y)}{p(z_y~|~t_x,t_y)}\bigg]dz_xdz_ydt_xdt_yds \nonumber
    \end{aligned}\\
    &= \begin{aligned}[t]
        &\mathbb{E}_{p(s,t_x,t_y,z_x)}\bigg[\int p(z_y~|~y)\log p(z_y~|~z_x, t_x, t_y)dz_y\bigg] \\
        &- \mathbb{E}_{p(s, t_x, t_y, z_x, z_y)}\bigg[\log p(z_y~|~t_x, t_y)\bigg] \nonumber
    \end{aligned}\\
    &= \begin{aligned}[t]
        &-\newunderbrace{\mathbb{E}_{p(s,t_x,t_y,z_x)}\bigg[H\Big(p(z_y~|~y), p(z_y~|~z_x,t_x,t_y)\Big)\bigg]}{\text{prediction term for target representation}}\quad \\
        &+ \newunderbrace{\mathbb{E}_{p(t_y)}\bigg[H\Big(p(z_y~|~t_y)\Big)\bigg]}{\text{regularization term to avoid collapse}},
    \end{aligned}
    \raisetag{30pt}
\end{align}
where the third equation holds because two representations are independent given the input and target, and in the last equation we apply the definitions of entropy and cross-entropy. Eq.~(\ref{equ:append_mutual_info}) shows that the mutual information can be divided into a prediction term and a regularization term. The prediction term requires the predicted distribution to be close to the target distribution, while the regularization term requires the target representations to maintain high entropy.

Next, we introduce parameterization to actually compute these terms. Two representations are encoded via an input encoder $f_\theta$ and a target encoder $f_\phi$, respectively. Because we do not know $p(z_y~|~z_x, t_x, t_y)$ in advance, we adopt an approximation by first predicting $\hat{z}_y=f_\psi(z_x, t_x, t_y)$ and then estimating with the posterior distribution $\hat{P}(z_y~|~\hat{z}_y)$. 
The mutual information thus becomes
% copy eqn 6/10
% \bm{z}
% eqn 12 split
\begin{align}
\small
\label{equ:appendix_LB_mutual_info}
    &I(z_x;z_y~|~t_x,t_y) \nonumber\\
    &= \begin{aligned}[t]
        \int p(t_x, t_y)\int\bigg[&p(z_x~|~t_x,t_y)p(z_y~|~z_x,t_x,t_y)\cdot \\
        &\log\frac{p(z_y~|~z_x,t_x,t_y)}{p(z_y~|~t_x,t_y)}\bigg]dz_xdz_ydt_xdt_y \nonumber
    \end{aligned}\\
    &=\begin{aligned}[t]
        &\mathbb{E}_{p(z_x,t_x,t_y)}\bigg[\int p(z_y~|~z_x, t_x, t_y)\log p(z_y~|~z_x, t_x, t_y)dz_y\bigg] \\
        &- \mathbb{E}_{p(z_x, z_y, t_x, t_y)}\bigg[\log p(z_y~|~t_x, t_y)\bigg] \nonumber
    \end{aligned}\\
    &=\begin{aligned}[t]
        &\newunderbrace{\mathbb{E}_{p(z_x,t_x,t_y)}\bigg[\int p(z_y~|~z_x, t_x, t_y)\log \frac{p(z_y~|~z_x, t_x, t_y)}{\hat{P}(z_y~|~\hat{z}_y)}dz_y\bigg]}{\text{KL Divergence $\geq 0$}} \\
        &+\mathbb{E}_{p(z_x,t_x,t_y)}\bigg[\int p(z_y~|~z_x, t_x, t_y)\log \hat{P}(z_y~|~\hat{z}_y)dz_y\bigg] \\
        &- \mathbb{E}_{p(z_x, z_y, t_x, t_y)}\bigg[\log p(z_y~|~t_x, t_y)\bigg] \nonumber
    \end{aligned}\\
    &\geq \begin{aligned}[t]
        &\mathbb{E}_{p(z_x,t_x,t_y)}\bigg[\int p(z_y~|~z_x, t_x, t_y)\log \hat{P}(z_y~|~\hat{z}_y)dz_y\bigg] \\
        &- \mathbb{E}_{p(z_x, z_y, t_x, t_y)}\bigg[\log p(z_y~|~t_x, t_y)\bigg] \nonumber
    \end{aligned}\\
    &= \begin{aligned}[t]
        &\begin{aligned}[t]
            \mathbb{E}_{p(z_x,z_y,t_x,t_y)}\bigg[\log \hat{P}(z_y~|~\hat{z}_y)\cdot
            \newunderbrace{\int p(s~|~z_x,z_y,t_x,t_y)ds}{\text{the integral is equal to $1$}}\bigg] 
        \end{aligned}\\
        &- \begin{aligned}[t]
            \mathbb{E}_{p(t_y)}\bigg[\int p(z_y~|~t_y)&\log p(z_y~|~t_y) \cdot \\
            &~~~\newunderbrace{\int p(z_x,t_x~|~z_y,t_y) dz_xdt_x}{this integral is equal to 1}dz_y \bigg] 
        \end{aligned}\nonumber
    \end{aligned}\\
    &= \begin{aligned}[t]
        &\newunderbrace{\mathbb{E}_{p(s,t_x,t_y)}\bigg[\log \hat{P}\Big(z_y(\phi)~|~\hat{z}_y(\theta,\psi)\Big)\bigg]}{\text{prediction term for target representation}} \qquad\qquad\qquad\\
        &+ \newunderbrace{\mathbb{E}_{p(t_y)}\bigg[H(z_y(\phi)~|~t_y)\bigg]}{\text{regularization term to avoid collapse}},
    \end{aligned}
    \raisetag{30pt}
\end{align}
where the fourth inequality holds because KL Divergence will not be less than $0$. In the fifth equality, we introduce training sample $s$ to the expectation of the first term and move $z_x$ and $t_x$ from the expectation of the second term. In the last equality, $z_x$ and $z_y$ is moved out of the expectation because they should be deterministic once $s$, $t_x$, $t_y$ and model parameters are given. The right-hand side of Eq.~(\ref{equ:appendix_LB_mutual_info}) is a lower bound of the actual mutual information and will be equal to it if and only if the estimated distribution $\hat{P}(z_y~|~\hat{z}_y)$ matches the real distribution $p(z_y~|~z_x, t_x, t_y)$. We note that because $z_y$ should be a deterministic feature given $z_x, t_x, t_y$ during training, equality can be achieved when the decoder predicts the target representation precisely. So we have
\smallalign{\begin{align}
\label{equ:append_sup_mutual_info}
    I(z_x; &z_y ~|~ t_x, t_y) = ~\sup_{f_\psi}~ \newunderbrace{\mathbb{E}_{p(t_y)} \Big[H\big(p(z_y(\phi) ~|~ t_y)\big) \Big]}{\text{regularization term to avoid collapse}} \nonumber\\
    &+\newunderbrace{\mathop{\mathbb{E}}_{p(s, t_x, t_y)}\Big[\log \hat{P}\big(z_y(\phi)~|~\hat{z}_y(\theta, \psi)\big)}{\text{prediction term for target representation}}\Big].
\end{align}}%
We usually deal with the regularization term in an implicit manner, such as introducing negative samples or stopping gradient to the target encoder. Therefore, the prediction term presents the loss function to be optimized in practice.

\subsection{Multi-input Multi-target Pre-training}
To derive the multi-input multi-target pre-training, we extend the input and the target to a set of $N$ inputs $X=\{x_i\}_{i=1}^N$ and $M$ targets $Y=\{y_j\}_{j=1}^M$. The set of targets are split into $K$ non-overlapping groups as $Y_m \cap Y_{n\ne m}=\varnothing, \cup_{k=1}^KY_k= Y$. The input representations and target representations are $\bm{z}_x=f_\theta(\{x_i\}_{i=1}^N)$ and $\bm{z}_y^k=f_{\phi_k}(Y_k)$, respectively. The mutual information is computed between $\bm{z}_x$ and $\{\bm{z}_y^k\}_{k=1}^K$ given the set of input transforms $\bm{t}_x$ and target transforms $\bm{t}_y$:
\begin{equation}
\small
    \max I(\bm{z}_x;\{\bm{z}_y^k\}_{k=1}^K~|~\bm{t}_x,\bm{t}_y).
\end{equation}
Similar to Eq.~(\ref{equ:append_mutual_info}), we can expand the mutual information as
\allowdisplaybreaks{
\begin{align}
    &I(\bm{z}_x;\{\bm{z}_y^k\}_{k=1}^K~|~\bm{t}_x,\bm{t}_y) \nonumber\\
    &= \begin{aligned}[t]
        \int p(\bm{t}_x, &\bm{t}_y)\int\bigg[p(\bm{z}_x,\{\bm{z}_y^k\}_{k=1}^K~|~\bm{t}_x,\bm{t}_y)\cdot \\
        &\log\frac{p(\{\bm{z}_y^k\}_{k=1}^K~|~\bm{z}_x,\bm{t}_x,\bm{t}_y)}{p(\{\bm{z}_y^k\}_{k=1}^K~|~\bm{t}_x,\bm{t}_y)}\bigg]d\bm{z}_xd\{\bm{z}_y^k\}_{k=1}^Kd\bm{t}_xd\bm{t}_y \nonumber
    \end{aligned}\\
    &= \begin{aligned}[t]
        \int p(&\bm{t}_x, \bm{t}_y)\int\bigg[p(\bm{z}_x,\{\bm{z}_y^k\}_{k=1}^K~|~s,\bm{t}_x,\bm{t}_y)p(s~|~\bm{t}_x,\bm{t}_y)\cdot \\
        &\log\frac{p(\{\bm{z}_y^k\}_{k=1}^K~|~\bm{z}_x,\bm{t}_x,\bm{t}_y)}{p(\{\bm{z}_y^k\}_{k=1}^K~|~\bm{t}_x,\bm{t}_y)}\bigg]d\bm{z}_xd\{\bm{z}_y^k\}_{k=1}^Kd\bm{t}_xd\bm{t}_yds \nonumber
    \end{aligned}\\
    &= \begin{aligned}[t]
        &\begin{aligned}[t]
            \mathbb{E}_{p(s,\bm{t}_x,\bm{t}_y,\bm{z}_x)}\bigg[\int p(&\{\bm{z}_y^k\}_{k=1}^K~|~Y)\cdot\\
            &\log p(\{\bm{z}_y^k\}_{k=1}^K~|~\bm{z}_x, \bm{t}_x, \bm{t}_y)d\{\bm{z}_y^k\}_{k=1}^K\bigg] 
        \end{aligned}\\
        &- \mathbb{E}_{p(s, \bm{t}_x, \bm{t}_y, \bm{z}_x, \{\bm{z}_y^k\}_{k=1}^K)}\bigg[\log p(\{\bm{z}_y^k\}_{k=1}^K~|~\bm{t}_x, \bm{t}_y)\bigg] \nonumber
    \end{aligned}\\
    &= \begin{aligned}[t]
        &\begin{aligned}[t]
            \sum_{k=1}^K&\mathbb{E}_{p(s,\bm{t}_x,\bm{t}_y,\bm{z}_x)}\bigg[\int p(\{\bm{z}_y^k\}_{k=1}^K~|~Y)\cdot\\
            &\qquad\qquad\log p(\bm{z}_y^k~|~\bm{z}_x, \bm{t}_x, \bm{t}_y, \{\bm{z}_y^i\}_{i=1}^{k-1})d\{\bm{z}_y^k\}_{k=1}^K\bigg] 
        \end{aligned}\\
        &- \mathbb{E}_{p(s, \bm{t}_x, \bm{t}_y, \bm{z}_x, \{\bm{z}_y^k\}_{k=1}^K)}\bigg[\log p(\{\bm{z}_y^k\}_{k=1}^K~|~\bm{t}_x, \bm{t}_y)\bigg]
    \end{aligned} \nonumber\\
    &= \begin{aligned}[t]
        &\begin{aligned}[t]
            \sum_{k=1}^K&\mathbb{E}_{p(s,\bm{t}_x,\bm{t}_y,\bm{z}_x)}\bigg[\int p(\{\bm{z}_y^i\}_{i=1}^{k-1}~|~Y)p(\bm{z}_y^k~|~Y)\cdot\\
            &\qquad\qquad\quad~~~\newunderbrace{\int p(\{\bm{z}_y^i\}_{i=k+1}^{K}~|~Y) d\{\bm{z}_y^i\}_{i=k+1}^K}{the integral is equal to $1$}\cdot\\
            &\qquad\qquad\log p(\bm{z}_y^k~|~\bm{z}_x, \bm{t}_x, \bm{t}_y, \{\bm{z}_y^i\}_{i=1}^{k-1})d\{\bm{z}_y^i\}_{i=1}^k\bigg] 
        \end{aligned}\\
        &- \mathbb{E}_{p(s, \bm{t}_x, \bm{t}_y, \bm{z}_x, \{\bm{z}_y^k\}_{k=1}^K)}\bigg[\log p(\{\bm{z}_y^k\}_{k=1}^K~|~\bm{t}_x, \bm{t}_y)\bigg]
    \end{aligned} \nonumber\\
    &= \begin{aligned}[t]
        &-\newunderbrace{\begin{aligned}[t]
            \sum_{k=1}^K&\mathbb{E}_{p(s,\bm{t}_x,\bm{t}_y,z_x,\{z_y^i\}_{i=1}^{k-1})}\bigg[\\
            &\qquad H\Big(p(\bm{z}_y^k~|~Y_k),p(\bm{z}_y^k~|~\bm{z}_x, \bm{t}_x, \bm{t}_y, \{\bm{z}_y^i\}_{i=1}^{k-1})\Big)\bigg]
        \end{aligned}}{\text{prediction term for target representations}} \\
        &+ \newunderbrace{\mathbb{E}_{p(\bm{t}_y)}\bigg[H\Big(p(\{z_y^k\}_{k=1}^K~|~\bm{t}_y)\Big)\bigg]}{\text{regularization term to avoid collapse}},
    \end{aligned}
    \raisetag{35pt}
\end{align}}%
where the fifth equality hold because the target representations are independent given targets, and $\{\bm{z}_y^i\}_{i=1}^{k-1}=\varnothing$ for $k=1$.

During parameterization, we adopt different predictions $\bm{\hat{z}}_y^k=f_{\psi_k}(\bm{z}_x, \bm{t}_x, \bm{t}_y)$ and distributions $\hat{P}_k(\bm{z}_y^k~|~\bm{\hat{z}}_y^k)$ for different target groups. Then the mutual information can be converted into
\allowdisplaybreaks{
\begin{align}
\small
    &I(\bm{z}_x;\{\bm{z}_y^k\}_{k=1}^K~|~\bm{t}_x,\bm{t}_y) \nonumber\\
    &= \begin{aligned}[t]
        \int p(&\bm{t}_x, \bm{t}_y)\int\bigg[p(\bm{z}_x~|~\bm{t}_x,\bm{t}_y)p(\{\bm{z}_y^k\}_{k=1}^K~|~\bm{z}_x,\bm{t}_x,\bm{t}_y)\cdot \\
        &~~\log\frac{p(\{\bm{z}_y^k\}_{k=1}^K~|~\bm{z}_x,\bm{t}_x,\bm{t}_y)}{p(\{\bm{z}_y^k\}_{k=1}^K~|~\bm{t}_x,\bm{t}_y)}\bigg]d\bm{z}_xd\{\bm{z}_y^k\}_{k=1}^Kd\bm{t}_xd\bm{t}_y \nonumber
    \end{aligned}\\
    &= \begin{aligned}[t]
        &\begin{aligned}[t]
            \mathbb{E}_{p(\bm{t}_x,\bm{t}_y,\bm{z}_x)}\bigg[\int p(&\{\bm{z}_y^k\}_{k=1}^K~|~\bm{z}_x,\bm{t}_x,\bm{t}_y)\cdot\\
            &\log p(\{\bm{z}_y^k\}_{k=1}^K~|~\bm{z}_x, \bm{t}_x, \bm{t}_y)d\{\bm{z}_y^k\}_{k=1}^K\bigg] 
        \end{aligned}\\
        &- \mathbb{E}_{p(\bm{t}_x, \bm{t}_y, \bm{z}_x, \{\bm{z}_y^k\}_{k=1}^K)}\bigg[\log p(\{\bm{z}_y^k\}_{k=1}^K~|~\bm{t}_x, \bm{t}_y)\bigg]
    \end{aligned}\nonumber\\
    &= \begin{aligned}[t]
        &\newunderbrace{\begin{aligned}[t]
            \sum_{k=1}^K\mathbb{E}_{p(\bm{t}_x,\bm{t}_y,\bm{z}_x,\{\bm{z}_y^i\}_{i=1}^{k-1})}\bigg[\int p(\bm{z}_y^k~|~\bm{z}_x, \bm{t}_x, \bm{t}_y, \{\bm{z}_y^i\}_{i=1}^{k-1})\cdot\\
            \log \frac{p(\bm{z}_y^k~|~\bm{z}_x, \bm{t}_x, \bm{t}_y, \{\bm{z}_y^i\}_{i=1}^{k-1})}{\hat{P}_k(\bm{z}_y^k~|~\bm{\hat{z}}_y^k)}d\bm{z}_y^k\bigg]
        \end{aligned}}{\text{KL Divergence $\geq 0$}} \nonumber \\
        &\begin{aligned}[t]
            +\sum_{k=1}^K\mathbb{E}_{p(\bm{t}_x,\bm{t}_y,\bm{z}_x,\{\bm{z}_y^i\}_{i=1}^{k-1})}\bigg[\int p(\bm{z}_y^k~|&~\bm{z}_x, \bm{t}_x, \bm{t}_y, \{\bm{z}_y^i\}_{i=1}^{k-1})\cdot\\
            &\log \hat{P}_k(\bm{z}_y^k~|~\bm{\hat{z}}_y^k)d\bm{z}_y^k\bigg]\quad
        \end{aligned} \nonumber \\
        &- \mathbb{E}_{p(\bm{t}_x, \bm{t}_y, \bm{z}_x, \{\bm{z}_y^k\}_{k=1}^K)}\bigg[\log p(\{\bm{z}_y^k\}_{k=1}^K~|~\bm{t}_x, \bm{t}_y)\bigg]
    \end{aligned}\nonumber\\
    &\geq \begin{aligned}[t]
        &\begin{aligned}[t]
            \sum_{k=1}^K\mathbb{E}_{p(\bm{t}_x,\bm{t}_y,\bm{z}_x,\{\bm{z}_y^i\}_{i=1}^{k-1})}\bigg[\int p(\bm{z}_y^k~|&~\bm{z}_x, \bm{t}_x, \bm{t}_y, \{\bm{z}_y^i\}_{i=1}^{k-1})\cdot\\
            &\log \hat{P}_k(\bm{z}_y^k~|~\bm{\hat{z}}_y^k)d\bm{z}_y^k\bigg]\quad
        \end{aligned} \\
        &- \mathbb{E}_{p(\bm{t}_x, \bm{t}_y, \bm{z}_x, \{\bm{z}_y^k\}_{k=1}^K)}\bigg[\log p(\{\bm{z}_y^k\}_{k=1}^K~|~\bm{t}_x, \bm{t}_y)\bigg]
    \end{aligned}\nonumber\\
    &= \begin{aligned}[t]
        &\begin{aligned}[t]
            \sum_{k=1}^K\mathbb{E}_{p(\bm{t}_x,\bm{t}_y,\bm{z}_x,\{\bm{z}_y^i\}_{i=1}^{k})}&\bigg[\log \hat{P}_k(\bm{z}_y^k~|~\bm{\hat{z}}_y^k)\cdot\\
            &\newunderbrace{\int p(s~|~\bm{t}_x,\bm{t}_y,\bm{z}_x,\{\bm{z}_y^i\}_{i=1}^{k})ds}{\text{the integral is equal to $1$}}\bigg]
        \end{aligned}\\
        &+\begin{aligned}[t]
            \mathbb{E}_{p(\bm{t}_y)}&\bigg[p(\{\bm{z}_y^k\}_{k=1}^K~|~\bm{t}_y)\log p(\{\bm{z}_y^k\}_{k=1}^K~|~\bm{t}_y)\cdot\\
            &~\newunderbrace{\int p(\bm{z}_x,\bm{t}_x~|~\{\bm{z}_y^k\}_{k=1}^K,\bm{t}_y)d\bm{z}_xd\bm{t}_x}{\text{the integral is equal to $1$}}d\{\bm{z}_y^k\}_{k=1}^K\bigg]
        \end{aligned}
    \end{aligned}\nonumber\\
    &= \begin{aligned}[t]
        &\newunderbrace{\sum_{k=1}^K\mathbb{E}_{p(s,\bm{t}_x,\bm{t}_y)}\bigg[\log \hat{P}_k\Big(\bm{z}_y^k(\phi_k)~|~\bm{\hat{z}}_y^k(\theta,\psi_k)\Big)\bigg]}{\text{prediction term for target representations}}\quad\quad\\
        &+\newunderbrace{\mathbb{E}_{p(\bm{t}_y)}\bigg[H(\{\bm{z}_y^k\}_{k=1}^K~|~ \bm{t}_y)\bigg]}{\text{regularization term to avoid collapse}},
    \end{aligned}
    \raisetag{30pt}
\end{align}}%
where the fourth inequality holds because KL Divergence will not be less than $0$ for every summation term. The equality can be achieved if and only if every $\hat{P}_k(\bm{z}_y^k~|~\bm{\hat{z}}_y^k)$ matches $p(\bm{z}_y^k~|~\bm{z}_x, \bm{t}_x, \bm{t}_y, \{\bm{z}_y^i\}_{i=1}^{k-1})$.
Therefore, the mutual information for multi-input multi-target pre-training can be bounded by
\smallalign{\begin{align}
\label{equ:multi_mutual}
    &I\big(\bm{z}_x; \{\bm{z}_y^k\}_{k=1}^{K}~|~\bm{t}_x, \bm{t}_y\big) \nonumber\\
    &\geq \sup_{\{f_{\psi_k}\}_{k=1}^{K}} \newunderbrace{\mathop{\mathbb{E}}_{p(\bm{t}_y)} \bigg[H\Big(p\big(\{\bm{z}_y^k(\phi_k)\}_{k=1}^{K}~|~\bm{t}_y\big)\Big) \bigg]}{regularization term to avoid collapse} \nonumber\\
    &+\newunderbrace{\sum_{k=1}^K~ \mathop{\mathbb{E}}_{p(\bm{s}, \bm{t}_x, \bm{t}_y)}\Big[\log \hat{P}_k\big(\bm{z}_y^k(\phi_k)~|~\bm{\hat{z}}_y^k(\theta,\psi_k)\big)\Big]}{prediction term for target representation}.
\end{align}}%
It's shown that different target groups are disentangled into a summation of prediction terms, so we can optimize each target objective independently.

\section{Experiment Details}

\subsection{Pre-training Settings}
\label{sec:appendix_pre}
We utilize InternImage-H as image encoder in Sec~\ref{sec:exp_large_model} for large model pre-training and ViT-B/16 as that in other experiments for ablation study and fair comparison. For image-text dataset (\eg YFCC-15M~\cite{thomee2016yfcc100m}), a 12-layer Transformer (with the same network architecture as BERT-Base~\cite{devlin2018bert}) is utilized as text target encoder. For image classification dataset (\eg ImageNet~\cite{deng2009imagenet}), we directly use the linear classifier weight as category embedding target. We employ 4-layer Transformer as decoder for image representation target, and Attention Pooling as that for category embedding or text global feature. Detailed hyper-parameters for pre-training InternImage-H and ViT-B are listed in Tab.~\ref{tab:hypers_pretrain}.

\vspace{0.5em}\noindent\textbf{Dynamic weighting} is used to balance the weights of self-supervised loss ($L_{\text{SSP}}$) and supervised/weakly-supervised loss ($L_{\text{SP}}$). The overall training loss can be expressed as
\begin{equation}
    L=L_{\text{SSP}} + \lambda L_{\text{SP}},
\label{equ:appendix_dw}
\end{equation}
where $\lambda$ is the balance loss weight. Because the loss behavior changes dramatically during training, it's sub-optimal to set a static weight. We propose to set $\lambda$ dynamically according to the loss gradients. Specifically, we compute the exponential moving average of gradient norm that each loss back-propagates to input features, denoted as $\bar{g}_{\text{uni-modal}}$ and $\bar{g}_{\text{multi-modal}}$. Then $\lambda$ is set as $ \gamma \cdot \bar{g}_{\text{uni-modal}}/\bar{g}_{\text{multi-modal}}$, where $\gamma$ controls the gradient ratio between two loss terms. We find this strategy to work well in practice ($\gamma = 1$ by default).

\begin{table}[h]
    \centering
    \small
    \resizebox{1.0\linewidth}{!}{
    \begin{tabular}{lcc}
    \toprule
        Hyper-parameters & ViT-B/16 & InternImage-H\\
    \midrule
        Image-to-image decoder layers & \multicolumn{2}{c}{4} \\
        Image-to-image decoder hidden size & 768 & 1024 \\
        Image-to-image decoder FFN hidden size & 3072 & 4096 \\
        Image-to-image decoder attention heads & \multicolumn{2}{c}{16} \\
        Attention pooling input size & 768 & 1024 \\
        Attention pooling output size & \multicolumn{2}{c}{768} \\
        Attention pooling attention heads & \multicolumn{2}{c}{16} \\
    \midrule
        Data augment & \multicolumn{2}{c}{\makecell{RandomResizedCrop\\RandomHorizontalFlip\\ColorJitter\\RandomGrayscale\\GaussianBlur\\Solarize}} \\
        Mask strategy & \multicolumn{2}{c}{Blockwise mask} \\
        Mask ratio & \multicolumn{2}{c}{50\%} \\
        Input resolution & $224\times 224$ & $192\times 192$ \\
    \midrule
        Training epochs & \makecell{1600(ImageNet)\\138(YFCC)} & 30 \\
        Batch size & 4096 & 40000 \\
        Adam $\beta$ & \multicolumn{2}{c}{(0.9, 0.95)} \\
        Peak learning rate & \multicolumn{2}{c}{$1.0\times 10^{-3}$} \\
        Learning rate schedule & \multicolumn{2}{c}{cosine} \\
        Warmup epochs & \makecell{40(ImageNet)\\3.5(YFCC)} &  1 \\
        Weight decay & \multicolumn{2}{c}{0.1} \\
        EMA coeff & \multicolumn{2}{c}{0.995} \\
        EMA schedule & \multicolumn{2}{c}{cosine} \\
        Label smoothing & \multicolumn{2}{c}{0.1} \\
        Stock. depth & 0.1 (linear) & 0.2 (uniform) \\
    \bottomrule
    \end{tabular}}
    \caption{Hyper-parameters for pre-training.}
    \label{tab:hypers_pretrain}
\end{table}

\subsection{Tranfer Settings of InternImage-H}

We strictly follow \cite{anonymous2022internimg} for the transfer settings of InternImage-H on ImageNet-1k, COCO, LVIS and ADE20k. We briefly summarize the settings below.

\vspace{0.5em}\noindent\textbf{ImageNet-1k.~} For ImageNet classification, the pre-trained InternImage-H is fine-tuned on ImageNet-1k for 30 epochs.

\vspace{0.5em}\noindent\textbf{COCO.~} For COCO object detection, we double the parameters of pre-trained InternImage-H via the composite techniques~\cite{liang2022cbnet}. Then it is fine-tuned with the DINO~\cite{zhang2022dino} detector on Objects365~\cite{shao2019objects365} and COCO datasets one after another for 26 epochs and 12 epochs.

\vspace{0.5em}\noindent\textbf{LVIS.~} For LVIS long-tailed object detection, we double the parameters of pre-trained InternImage-H via the composite techniques~\cite{liang2022cbnet}. Then it is fine-tuned with the DINO~\cite{zhang2022dino} detector on Objects365~\cite{shao2019objects365} and LVIS datasets one after another for 26 epochs and 12 epochs.

\vspace{0.5em}\noindent\textbf{ADE20k.~} For ADE20k semantic segmentation, we fine-tune InternImage-H with Mask2Former~\cite{cheng2021masked}, and adopt the same settings in \cite{wang2022image,chen2022vitadapter}. 

\subsection{Transfer Settings of ViT-B/16}

\noindent\textbf{ImageNet-1k.~} The detailed fine-tuning and linear classification settings of ViT-B/16 on ImageNet-1k are listed in Tab.~\ref{tab:hypers_finetune} and Tab.~\ref{tab:hypers_linear}. 

\vspace{0.5em}\noindent\textbf{COCO and LVIS.~} We utilize ViTDet~\cite{li2022exploring} for object detection. By default, the fine-tuning schedule is set to 100 epochs for both COCO and LVIS datasets. For the ablation study, we use a short schedule of 25 epochs. Detailed hyper-parameters are listed in Tab.~\ref{tab:hypers_coco} and Tab.~\ref{tab:hypers_lvis}.

\vspace{0.5em}\noindent\textbf{ADE20k.~} Following~\cite{bao2021beit,he2022masked,anonymous2022siamese}, we employ UperNet~\cite{xiao2018unified} as the segmentation network. We use the implementation in MMSegmentation~\cite{mmseg2020}. Detailed hyper-parameters are listed in Tab.~\ref{tab:hypers_seg}.

\begin{table}[h]
    \centering
    \small
    \begin{tabular}{lc}
    \toprule
        Hyper-parameters & Value \\
    \midrule
        Erasing prob. & 0.25 \\
        Rand augment & 9/0.5 \\
        Mixup prob. & 0.8 \\
        Cutmix prob. & 1.0 \\
        Input resolution & $224\times 224$ \\
    \midrule
        Finetuning epochs & 100 \\
        Batch size & 1024 \\
        Adam $\beta$ & (0.9, 0.999) \\
        Peak learning rate & $2.0\times 10^{-3}$\\
        Learning rate schedule & cosine \\
        Warmup epochs & 5 \\
        Weight decay & 0.1 \\
        Layer-wise learning rate decay & 0.65 \\
        Label smoothing & 0.1 \\
        Stock. depth & 0.1 \\
    \bottomrule
    \end{tabular}
    \caption{Hyper-parameters of ViT-B for ImageNet finetuning.}
    \label{tab:hypers_finetune}
\end{table}
\begin{table}[h]
    \centering
    \small
    \begin{tabular}{lc}
    \toprule
        Hyper-parameters & Value \\
    \midrule
        Data augment & \makecell{RandomResizedCrop\\RandomHorizontalFlip} \\
        Input resolution & $224\times 224$ \\
    \midrule
        Training epochs & 90 \\
        Batch size & 16384 \\
        Optimizer & LARS \\
        Peak learning rate & 3.2 \\
        Learning rate schedule & cosine \\
        Warmup epochs & 10 \\
        Weight decay & 0.0 \\
    \bottomrule
    \end{tabular}
    \caption{Hyper-parameters of ViT-B for ImageNet linear probing.}
    \label{tab:hypers_linear}
\end{table}
\begin{table}[h]
    \centering
    \small
    \begin{tabular}{lc}
    \toprule
        Hyper-parameters & Value \\
    \midrule
        Data augment & large scale jittor \\
        Input resolution & $1024\times 1024$ \\
    \midrule
        Finetuning epochs & 100 \\
        Batch size & 64 \\
        Adam $\beta$ & (0.9, 0.999) \\
        Peak learning rate & $1.0\times 10^{-4}$\\
        Learning rate schedule & step \\
        Warmup length & 250 iters \\
        Weight decay & 0.1 \\
    \midrule
        Stock. depth & 0.1 \\
    \midrule
        Relative positional embeddings & \checkmark \\
    \bottomrule
    \end{tabular}
    \caption{Hyper-parameters of ViT-B for COCO detection.}
    \label{tab:hypers_coco}
\end{table}
\begin{table}[h]
    \centering
    \small
    \begin{tabular}{lc}
    \toprule
        Hyper-parameters & Value \\
    \midrule
        Data augment & large scale jittor \\
        Input resolution & $1024\times 1024$ \\
    \midrule
        Finetuning epochs & 100 \\
        Batch size & 64 \\
        Adam $\beta$ & (0.9, 0.999) \\
        Peak learning rate & $2.0\times 10^{-4}$\\
        Learning rate schedule & step \\
        Warmup length & 250 iters \\
        Weight decay & 0.1 \\
    \midrule
        Stock. depth & 0.1 \\
    \midrule
        Relative positional embeddings & \checkmark \\
    \bottomrule
    \end{tabular}
    \caption{Hyper-parameters of ViT-B for LVIS detection.}
    \label{tab:hypers_lvis}
\end{table}
\begin{table}[h]
    \centering
    \small    
    \begin{tabular}{lc}
    \toprule
        Hyper-parameters & Value \\
    \midrule
        Data augment & \makecell{RandomCrop\\RandomFlip\\PhotoMetricDistortion} \\
        Input resolution & $512\times 512$ \\
    \midrule
        Finetuning length & 160k iters \\
        Batch size & 16 \\
        Adam $\beta$ & (0.9, 0.999) \\
        Peak learning rate & $1.0\times 10^{-4}$\\
        Learning rate schedule & linear \\
        Warmup length & 1500 iters \\
        Weight decay & 0.05 \\
    \midrule
        Stock. depth & 0.1 \\
    \midrule
        Relative positional embeddings & \checkmark \\
    \bottomrule
    \end{tabular}
    \caption{Hyper-parameters of ViT-B for ADE20k semantic segmentatioin.}
    \label{tab:hypers_seg}
\end{table}

\subsection{More Experiments}

\setlength{\tabcolsep}{4pt}
\begin{table}
    \centering
    \small
    \resizebox{0.8\linewidth}{!}{
        \begin{tabular}{cccccc}
        \Xhline{2\arrayrulewidth}
        Gradient Ratio $\gamma$ & 0.2 & 0.5 & 1.0 & 2.0 & 5.0 \\
        \hline
        ImageNet Top1 & 83.1 & 83.2 & \default{83.3} & 82.8 & 82.5 \\
        COCO $\text{AP}^\text{box}$ & 50.2 & \default{50.5} & \default{50.5} & 48.9 & 47.6\\
        \Xhline{2\arrayrulewidth}
        \end{tabular}
    }
    \vspace{-1em}
    \caption{Ablation study of gradient ratio $\gamma$.}
    \vspace{-1.5em}
    \label{tab:exp_ablation_grad_ratio}
\end{table}

\noindent\textbf{Ablation Study on Gradient Ratio $\gamma$.~} The gradient ratio $\gamma$ is used in dynamic weighting (see Eq.~\eqref{equ:appendix_dw} in Appendix~\ref{sec:appendix_pre}). We ablate the choice of $\gamma$ from $\{0.2, 0.5, 1.0, 2.0, 5.0\}$ in Tab.~\ref{tab:exp_ablation_grad_ratio}. These models are pre-trained on ImageNet-1k for 100 epochs. Then they are fine-tuned on ImageNet-1k classification and COCO object detection. The fine-tuning schedules for ImageNet-1k and COCO are set to 100 epochs and 25 epochs respectively. As shown in Tab.~\ref{tab:exp_ablation_grad_ratio}, $\gamma=0.2, 0.5, 1.0$ works quite well in both classification and detection. We choose $\gamma=1.0$ as our default setting for its simplicity.

\begin{figure}
    \centering
    \includegraphics[width=0.95\linewidth]{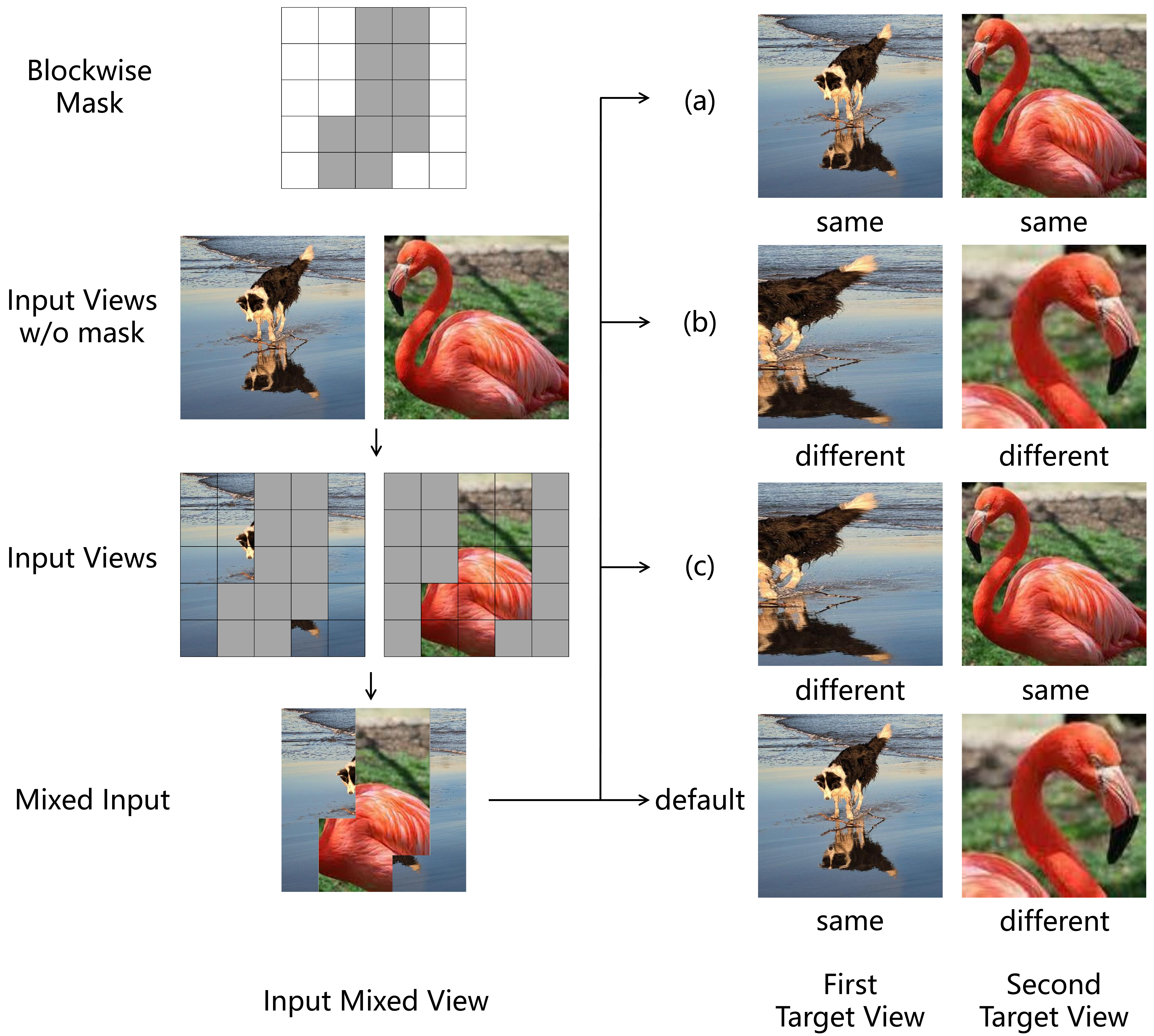}
    \vspace{-0.5em}
    \caption{Illustration of four design choices of target views. }
    \label{fig:target_views}
\end{figure}

\setlength{\tabcolsep}{4pt}
\begin{table}
    \centering
    \small
    \resizebox{1.0\linewidth}{!}{
        \begin{tabular}{ccccc}
        \Xhline{2\arrayrulewidth}
        & First Target View & Second Target View & ImageNet Top1$^{\dag}$ & COCO \\ 
        \hline
        (a) & same      & same        & 77.2 & 48.6 \\
        (b) & different & different   & 78.5 & \default{49.8} \\
        (c) & different      & same   & 78.8 & 49.2 \\
        \default{default} & \default{same} & \default{different}  & \default{79.1} & 49.5 \\
        \Xhline{2\arrayrulewidth}
        \end{tabular}
    }
    % \vspace{-1em}
    \caption{Ablation study of target views. $^{\dag}$ ImageNet fine-tuning is early-stopped at 20 epochs which we found consistent with the final performance in practice.}
    \vspace{-1.5em}
    \label{tab:exp_ablation_target_views}
\end{table}

\vspace{0.5em}\noindent\textbf{Ablation Study on Target Views.~} Our \name{} consists of two target image views during the multi-input multi-target pre-training. Two input views of different images are mixed with a shared blockwise mask. As shown in Fig.~\ref{fig:target_views}, the visible part of the blockwise mask is filled with an augmented view of the first image, and the masked part is filled with an augmented view of the second image. The first target view and second target view are not permutable. We ablate the choices of these two target image views (either the same or different from the input image view) in Tab.~\ref{tab:exp_ablation_target_views}. These models are pre-trained on ImageNet-1k without labels (\ie they only have image representation target and do not have the category embedding target) for 200 epochs. Then, they are fine-tuned on ImageNet-1k classification and COCO object detection. The fine-tuning schedule is set to 100 epochs and 25 epochs respectively. Our default setting works best in ImageNet classification. Although (b) perform slightly better than our default setting in COCO detection, the pre-training process of it is quite unstable (FP16 loss scale is quite unstable), thus we do not choose it as our default setting.

\end{document}